\documentclass{article}

\usepackage[preprint, nonatbib]{neurips_2025}

\usepackage[utf8]{inputenc} %
\usepackage[T1]{fontenc}    %
\usepackage{hyperref}       %
\usepackage{url}            %
\usepackage{booktabs}       %
\usepackage{amsfonts}       %
\usepackage{nicefrac}       %
\usepackage{microtype}      %
\usepackage{xcolor}         %
\usepackage{microtype}
\usepackage{graphicx}
\usepackage{booktabs} %
\usepackage{multirow}
\usepackage{subcaption}
\usepackage{algorithm}
\usepackage{algpseudocode}
\usepackage{amsmath}
\usepackage{amssymb}
\usepackage{mathtools}
\usepackage{amsthm}
\usepackage[numbers,compress]{natbib}
\usepackage[capitalize,noabbrev]{cleveref}

\title{TW-CRL: Time-Weighted Contrastive Reward Learning for Efficient Inverse Reinforcement Learning}

\newcommand{\sgtrap}{s_{\text{trap}}}
\newcommand{\Sgtrap}{\mathcal{S}_{\text{trap}}}
\newcommand{\Sgoal}{\mathcal{S}_{\text{goal}}}
\newcommand{\sgoal}{s_{\text{goal}}}
\newcommand{\Pgoal}{P(s_t \in \Sgoal)}
\newcommand{\Ptrap}{P(s_t \in \Sgtrap)}
\newcommand{\StateS}{\mathcal{S}}
\newcommand{\Action}{\mathcal{A}}
\newcommand{\Horizon}{T}
\newcommand{\Dynamics}{P}

\author{
	Yuxuan Li$^{\dagger}$ \quad Yicheng Gao$^{\dagger}$ \quad Ning Yang$^\diamondsuit$ \quad Stephen Xia$^{\dagger}$ \\
	$^{\dagger}$ Northwestern University \\
	$^\diamondsuit$ Institute of Automation, Chinese Academy of Sciences \\
	\texttt{yuxuanli2023@u.northwestern.edu},  \texttt{ning.yang@ia.ac.cn} \\
}

\begin{document}

\theoremstyle{plain}
\newtheorem{theorem}{Theorem}[section]
\newtheorem{proposition}[theorem]{Proposition}
\newtheorem{lemma}[theorem]{Lemma}
\newtheorem{corollary}[theorem]{Corollary}
\theoremstyle{definition}
\newtheorem{definition}[theorem]{Definition}
\newtheorem{assumption}[theorem]{Assumption}
\theoremstyle{remark}
\newtheorem{remark}[theorem]{Remark}

\maketitle

\begin{abstract}
Episodic tasks in Reinforcement Learning (RL) often pose challenges due to sparse reward signals and high-dimensional state spaces, which hinder efficient learning. Additionally, these tasks often feature hidden ``trap states''—irreversible failures that prevent task completion but do not provide explicit negative rewards to guide agents away from repeated errors. To address these issues, we propose Time-Weighted Contrastive Reward Learning (TW-CRL), an Inverse Reinforcement Learning (IRL) framework that leverages both successful and failed demonstrations. By incorporating temporal information, TW-CRL learns a dense reward function that identifies critical states associated with success or failure. This approach not only enables agents to avoid trap states but also encourages meaningful exploration beyond simple imitation of expert trajectories. Empirical evaluations on eight benchmarks demonstrate that TW-CRL surpasses state-of-the-art methods, achieving improved efficiency and robustness. 
\end{abstract}

\section{Introduction}
An episodic task is a common Reinforcement Learning (RL) problem that has a well-defined termination condition~\cite{Sutton1999BetweenMA}. Many real-world problems, such as robotic manipulation and navigation, are episodic by nature, often featuring complex state and action spaces alongside sparse reward signals \cite{andrychowicz2018hindsightexperiencereplay,robosuite2020,yu2019meta}. The sparsity of rewards significantly complicates the training process by providing a positive reward only when the goal is achieved, leaving most steps uninformative. As a result, agents may spend excessive time in states that provide no meaningful feedback, slowing down the learning process. Moreover, certain “trap states” can make the goal permanently unreachable after a single mistake. These states do not explicitly offer negative rewards, but act as hidden traps that derail the agent’s attempts to succeed.

Designing dense reward functions is an effective approach to address challenges in RL environments with sparse rewards \cite{gehring2022reinforcementlearningclassicalplanning, Sutton2011HordeAS,IT1,IT2}. However, how to develop such reward functions remains a challenging task. A straightforward approach is to use distance-based rewards, where the agent receives rewards based on its distance from the goal \cite{trott2019keepingdistancesolvingsparse}. While this method is intuitive, distance-based reward may not be effective under more complex environments where the distance between current state and goal state is difficult to measure. Another promising approach to providing dense rewards is Inverse Reinforcement Learning (IRL) \cite{IRL,Abbeel2010InverseRL}, which infer reward functions from expert demonstrations. These methods assign higher rewards to states or state-action pairs that more closely align with expert behavior; however, this imitation-based strategy can limit exploration and cause agents to converge to suboptimal behaviors \cite{pmlr-v162-mavor-parker22a}. As a result, agents may fail to discover alternative strategies, such as more efficient paths or ways to avoid trap states (\cref{subsec:trapstates}). Additionally, most IRL-based approaches rely solely on successful demonstrations, which reduces data efficiency, especially in complex environments where failures are common during early training \cite{Shiarlis2016InverseRL}. By ignoring failed experiences, these methods miss valuable information that could help agents learn to avoid undesirable trajectories, ultimately slowing down the learning process and reducing the robustness of the learned policies.

To address these challenges, we propose Time-Weighted Contrastive Reward Learning (TW-CRL), a novel IRL approach that trains the agent based on a learned accurate and environment-aligned dense reward function. TW-CRL contains two key components: Time-Weighted function and Contrastive Reward Learning. The Time-Weighted function incorporates temporal information from trajectories, providing denser and more informative feedback. This allows the agent to receive more accurate rewards that reflect the importance of different states over time. Contrastive Reward Learning utilizes both successful and failed demonstrations, which enables the agent to efficiently locate goals while avoiding repeatedly falling into the same failure patterns, ultimately improving learning efficiency and average episodic return.

The key contributions of this paper are as follows: (1) We define and formalize the trap state problem in episodic RL tasks, highlighting its impact on agent performance. (2) We introduce TW-CRL, a novel IRL method that effectively learns dense and accurate reward functions from both successful and failed demonstrations.

\section{Related Work}
\textbf{IRL Approaches}
Most existing IRL methods focus primarily on imitating expert behavior. Maximum Entropy IRL (MaxEnt) \cite{Ziebart2008MaximumEI} encourages diverse actions while maintaining consistency with expert demonstrations by maximizing entropy. Generative Adversarial Imitation Learning (GAIL) \cite{ho2016generativeadversarialimitationlearning} and Adversarial Inverse Reinforcement Learning (AIRL) \cite{Fu2017LearningRR} applies adversarial learning to align the agent’s state-action distribution with that of the expert. Although recent research \cite{ICRL, IRL2, MERIT} continues to advance these methods, they often overlook the valuable insights that can be gained from failed demonstrations. As a result, their ability to recognize and avoid potential traps in complex environments is limited. TW-CRL overcomes this challenge by leveraging both successful and failed demonstrations, allowing agents to explore beyond simple imitation.

\textbf{Learning from Failures in RL}
Recent studies have explored leveraging failures to improve learning. Inverse Reinforcement Learning from Failure (IRLF) \cite{Shiarlis2016InverseRL} helps agents imitate expert behavior while avoiding past failures but does not fully analyze failure patterns. Self-Adaptive Reward Shaping (SASR) \cite{ma2024highlyefficientselfadaptivereward} adjusts rewards based on past success rates to balance exploration and exploitation. Other approaches such as \cite{chen2020learningsuboptimaldemonstrationselfsupervised, SQIL,vecerik2018leveragingdemonstrationsdeepreinforcement} and \cite{ICRL} incorporate suboptimal demonstrations to aid training but typically rely only on successful yet imperfect trajectories, ignoring completely failed ones. \cite{FL3, FL2, FL1, FL4} also focus on failure utilization but overlook the timing and sequence of failures. Unlike these methods, TW-CRL incorporates temporal information from both successful and failed demonstrations, leading to a more informative reward function.

\textbf{Risk Awareness in RL}
Risk-aware methods in RL focus on training agents to avoid risky situations \cite{Garca2015ACS}. Approaches such as Risk-Averse Imitation Learning (RAIL) \cite{santara2017railriskaverseimitationlearning,RA5,RA6} and Risk-Sensitive Generative Adversarial Imitation Learning (RS-GAIL) \cite{lacotte2018risksensitivegenerativeadversarialimitation} use the Conditional Value at Risk (CVaR) metric to assess and manage risk, while Inverse Risk-Sensitive Reinforcement Learning (IRSRL) \cite{ratliff2017inverserisksensitivereinforcementlearning} employs convex risk measures to minimize potential dangers. Recent studies \cite{RA1, RA2, RA3, RA4} also follow similar strategies by relying on predefined risk measures based on expert knowledge or historical data. In contrast, TW-CRL automatically identifies and avoids risky regions without requiring prior knowledge or explicit risk metrics.

\section{Problem Setting}\label{sec:problem}

\subsection{Episodic Tasks} 
An episodic MDP \cite{Sutton1999BetweenMA, Domingues2020EpisodicRL, Neu2020AUV} is formally defined as a tuple \( ({\mathcal{S}}, \Action, \Horizon, \mu, \Dynamics, r)\)\label{episodicMDP}, where ${\mathcal{S}}$ is the state space, $\Action$ is the action space, $\Horizon$ is the number of steps in one episode, $\mu$ is the distribution of the initial state, $\Dynamics: {\mathcal{S}} \times \Action \rightarrow {\mathcal{S}}$ is the dynamics which gives the distribution of the next state $s'$ given the current state $s$ and the action $a$, and $r$ is the reward function given by the environment. Episodic tasks, modeled by episodic MDPs, have a well-defined termination condition. In contrast, continuous tasks are tasks without clear termination conditions and proceeds indefinitely.

\subsection{Trap States}\label{subsec:trapstates}
Consider an episodic task with state space $\mathcal{S}$. For now, we assume that the task has a clear set of successful terminal states, meaning that there exists a non-empty set of goal states $\Sgoal\subseteq\mathcal{S}$ that contains all terminal states indicating successful completion. Once the agent reaches a goal state $\sgoal\in\Sgoal$, it remains in the same state until the end of that episode. We'll extend our problem setting to other task types in \cref{subsec:extend}. 

Given these assumptions, we define the concept of trap states. In episodic tasks, the agent may reach a state where it is impossible to reach a goal state, no matter what actions it takes. For example, in an object-pushing task, if the robot accidentally pushes an object off the table, it cannot recover and complete the task. These states are typically not explicitly defined or labeled in the environment and do not provide negative rewards to signal the agent’s mistakes. They act as unseen traps that impede the agent’s progress. We aim to create a reward function that gives negative feedback for these hidden trap states, helping the agent learn to avoid them during training. Concretely, we define the trap states as follows:
\begin{definition} [Trap state]
A trap state is a state $\sgtrap \in {\mathcal{S}}  $ such that, if $\sgtrap$ is a trap state, then regardless of action $a$, the next state $\sgtrap'$ is always a trap state. Additionally, all goal states are not trap states. Formally,

\begin{equation}
\Sgtrap = \{\sgtrap \in {\mathcal{S}} - \mathcal{S}_{\text{goal}} \mid \sum\limits_{s' \in \Sgtrap} P( s' \mid \sgtrap, a) = 1, \forall a \in \Action \}.
\end{equation}

\end{definition}
Here,  $\sgtrap$  is a trap state,  $\Sgtrap$  is the set of all trap states, $a$ is any action in the action space  $\Action$ , and  $s'$  is the next state determined by the environment based on  $\sgtrap$. This means once the agent enters such a state, it remains stuck in a set of trap states, as every action it takes keeps it in the same situation. 

\subsection{Generalization of Trap States} 
We then extend the definition of trap states to a broader situation, which applies to most cases—any situation where the agent fails to achieve the goal. Episodic tasks often have sparse rewards, meaning the agent only receives a positive reward upon reaching the goal and does not get helpful feedback during the rest of the process. As a result, the agent may spend a lot of time in states where it receives no meaningful information, making it harder to learn and improve, eventually leading to failure. Just like we want to give negative feedback for trap states, we also want to provide negative feedback for these uninformative states that contribute to failure. This can help the agent avoid repeating the same mistakes in future exploration and learning, improving its overall training efficiency.

Based on this idea, we consider all failed trajectories as a gradual process in which the agent moves toward a trap state. In other words, we assume that the final state of every failed trajectory is a trap state. Formally, in an episodic task, a failed trajectory $\tau^-$ is defined as a sequence of state-action pairs:

\begin{equation}
    \tau^- = \{(s_1^-, a_1^-),(s_2^-, a_2^-), \ldots , (s_T^-, a_T^-)\}.
\end{equation}

where $s_T^-\in\Sgtrap$. Contrarily, in an episodic task, the final state of a successful trajectory is always a goal state. Therefore, a successful trajectory $\tau^+$ can be represented similarly as:

\begin{equation}
    \tau^+=\{(s_1^+, a_1^+),(s_2^+, a_2^+), \ldots , (s_T^+, a_T^+)\}.
\end{equation}

where $s_T^+\in\Sgoal$. Furthermore, we can apply the notion above to a set of expert demonstrations $\mathcal{D}$. We denote the set of all successful demonstrations as $\mathcal{D}^+= \{ \tau^+ \in \mathcal{D}| s_T^+\in\StateS_{\text{goal}}\}$. The set of all failed demonstrations is denoted as $\mathcal{D}^- = \{\tau^- \in \mathcal{D}| s_T^-\in\Sgtrap\}$.

Based on these assumptions, we propose a method to generate a meaningful reward function that can help the agent learn more effectively.

\section{Method}
In this section, we introduce TW-CRL, a novel IRL method that utilizes temporal information from both failed and successful demonstrations in episodic tasks. We first describe how we use the Time-Weighted function to extract temporal information from trajectories (\cref{TW}). Next, we detail the use of Contrastive Reward Learning loss function to train a reward function that helps the agent identify unseen trap states and goal states (\cref{sec:CRL}). Then, we outline the entire TW-CRL process (\cref{subsec:pipeline}). Finally, we discuss how TW-CRL could be extended to tasks types beyond episodic tasks with clear sets of successful terminal states (\cref{subsec:extend}).

\begin{figure*}[tbh!]
    \centering
    \includegraphics[width=1\linewidth]{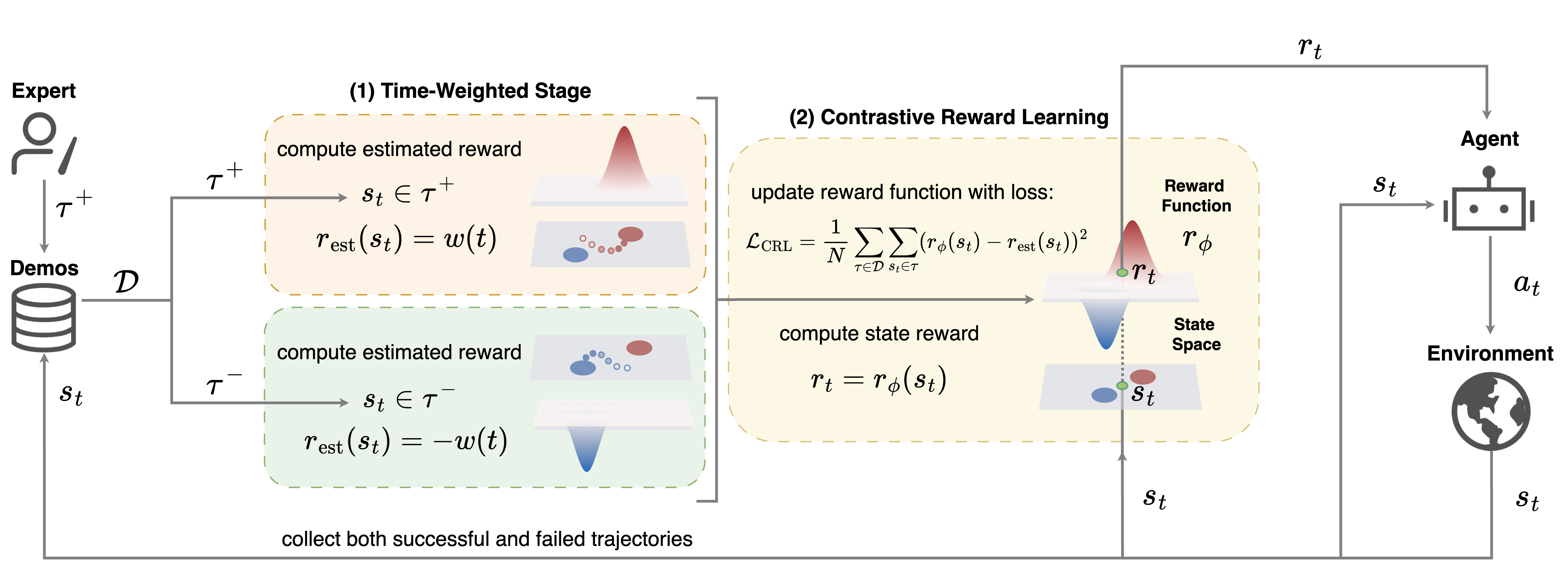}
    \caption{Overview of TW-CRL.}
    \label{fig:method}
\end{figure*}

\subsection{Time-Weighted Function}\label{TW}
We first describe how TW-CRL extracts temporal information from demonstrations. As discussed in \cref{subsec:trapstates}, in successful demonstrations, the agent gradually moves closer to the goal as the episode progresses, making states that appear later in the trajectory more likely to be goal states. Conversely, in failed demonstrations, the agent eventually reaches a trap state that prevents task completion, meaning that states appearing later in the trajectory are more likely to be trap states. To capture this temporal feature, we model the dynamics of the agent as an absorbing Markov chain. For failed demonstrations, the trap states are absorbing states and non-trap states are transient states. In a failed demonstration, recall that we assume the final state is always a trap state ($s_T\in\Sgtrap$). Additionally, we also assume that the initial state is not a trap state ($s_0\notin\Sgtrap$). Under these assumptions, given the prior knowledge that a demonstration is a failed demonstration, we can derive the conditional probability of a state $s_t$ of timestep $t$ being a trap state as follows:
\begin{align}\label{eq:pst}
    P(s_t\in\Sgtrap | s_T\in\Sgtrap) &= \frac{1-(1-f(t, T))^t}{1-(1-f(t, T))^T}
\end{align}
where $f:\mathbb{N}\times\mathbb{N}\rightarrow\mathbb{R}$ is a function that indicates the probability of a non-trap state transitioning into a trap state in timestep $t$ given total horizon $T$. The derivation of \cref{eq:pst} is in Appendix~\ref{ap:der-1}. In practice, $f(t, T)$ can be adjusted to meet different needs of specific environments. In our experiments, we let $f(t, T) \propto e^{\alpha t}$, where $\alpha > 0$ is a hyperparameter. This is because we observe an exponential increase with respect to the timestep matches the best with our environments. After normalization to ensure that $f(0, T) = 0$ and $f(T, T) = 1$, we have
\begin{align}\label{eq:ftT}
f(t, T) = \frac{e^{\alpha t} -1 }{e^{\alpha T} -1}
\end{align}
Given this assumption, \cref{eq:pst} can be further simplified:

\begin{align}\label{eq:wt-f}
P(s_t\in\Sgtrap | s_T\in\Sgtrap) \approx 1-(1-\frac{e^{\alpha t} -1 }{e^{\alpha T} -1})^t
\end{align}

For successful demonstrations, the goal states are absorbing states, and non-goal states are transient states. Similar to our conclusion above, we can derive that:

\begin{align}\label{eq:wt-s}
P(s_t\in\Sgoal | s_T\in\Sgoal) \approx 1-(1-\frac{e^{\alpha t} -1 }{e^{\alpha T} -1})^t
\end{align}

The derivations of \cref{eq:wt-f} and (\ref{eq:wt-s}) are in Appendix~\ref{ap:der-23}. We can observe that the right hand side of \cref{eq:wt-f} and (\ref{eq:wt-s}) are the same. We thus use it as our Time-Weighted function $w(t)$ for both successful and failed demonstrations:

\begin{align}\label{eq:wt}
    w(t) = 1-(1-\frac{e^{\alpha t} -1 }{e^{\alpha T} -1})^t
\end{align}

Further discussions regarding the properties of $w(t)$ is in Appendix~\ref{ap:wt}. As we'll discuss in \cref{sec:CRL}, the Time-Weighted function is used as a weighting function for states in different timesteps within each demonstration. By adopting the simplified conditional probabilities derived above as our Time-Weighted function, we make sure that higher weights are assigned to states that have heavier impacts on the success or failure of the trajectory they're in. Specifically, the weight assigned to each state equals the probability of the agent entering $\Sgoal$ / $\Sgtrap$ in that timestep.

\subsection{Contrastive Reward Learning}\label{sec:CRL}
In TW-CRL, the reward function  $r_{\phi}$  is modeled using a deep neural network and trained in a supervised manner. The training process involves creating a dataset from both successful and failed demonstrations. Concretely, successful and failed demonstrations are converted into supervised learning labels using the Time-Weighted function introduced previously. The key idea is to frame the estimation of  $r_{\phi}(s_t)$  as a supervised regression problem with contrastive labels, where successful demonstrations serve as “positive” examples and failed demonstrations as “negative” examples. 

Intuitively, we aim to train a reward function with labels $r_{\text{est}}(s_t)$ of the following form for each state $s_t$:
\begin{equation} \label{eqn:ideal}
   r_{\text{est}}(s_t) \propto \Pgoal - \Ptrap, 
\end{equation}

However, in continuous environments, it is difficult to obtain the same state in both successful and failed demonstrations to estimate $\Pgoal$ and  $\Ptrap$ for labeling. To work around this issue, we instead use the Time-Weighted function  $w(t)$ introduced in \cref{TW}  to compute the estimated reward  $r_{\text{est}}$  for each state as follows:
\begin{equation} \label{eqn:assign}
r_{\text{est}}(s_t) =
\begin{cases} 
w(t), & \text{if } s_t \in \tau^+ \\
-w(t), & \text{if } s_t \in \tau^- 
\end{cases}
\end{equation} 

For a state $s_t$  in a successful demonstration  $\tau^+$, we assign a positive value $w(t)$ as its estimated reward, meaning the state is more likely to help achieve the goal. For a state $ s_t $ in a failed demonstration  $\tau^- $, we assign a negative reward $-w(t)$, meaning the state is more likely to cause failure. We show in Appendix~\ref{ap:crl} that training the reward function using labels shown in \cref{eqn:assign} is similar to \cref{eqn:ideal}. We then introduce the \textbf{Contrastive Reward Learning loss function} to train the reward function in a simple way:
\begin{align}
\mathcal{L}_{\text{CRL}} &= 
 \frac{1}{N} \sum_{\tau \in \mathcal{D}} \sum_{s_t \in \tau} \left(r_{\phi}(s_t) - r_{\text{est}}(s_t)\right)^2 \\
&= \frac{1}{N} (\sum_{\tau \in \mathcal{D}^+} \sum_{s_t \in \tau} \left(r_{\phi}(s_t) - r_{\text{est}}(s_t)\right)^2  + \sum_{\tau \in \mathcal{D}^-} \sum_{s_t \in \tau} \left(r_{\phi}(s_t) - r_{\text{est}}(s_t)\right)^2) \\
&=\frac{1}{N} (\sum_{\tau \in \mathcal{D}^+} \sum_{s_t \in \tau} \left(r_{\phi}(s_t) - w(t)\right)^2  + \sum_{\tau \in \mathcal{D}^-} \sum_{s_t \in \tau} \left(r_{\phi}(s_t) + w(t)\right)^2),\label{CRL}
\end{align}
where $ \mathcal{D} = \mathcal{D}^+ \cup \mathcal{D}^- $ represents the collection of all trajectories from both successful and failed demonstrations, and $N$ is the total number of states. This loss function minimizes the Mean Squared Error (MSE) between the predicted rewards and the estimated reward of $s_t$. By minimizing the proposed loss function, the reward function is able to effectively learn to distinguish between states associated with successful task completion and those leading to failure.

\subsection{TW-CRL: Complete Algorithm Pipeline}\label{subsec:pipeline}

An overview of the TW-CRL framework is presented in \cref{fig:method}, and the corresponding pseudocode is provided in \cref{alg:tw-crl}. TW-CRL follows an iterative process that consists of training a reward function, optimizing a policy, and collecting new data through agent-environment interaction. Similar to other IRL methods, TW-CRL requires only successful demonstrations at the beginning, while failed demonstrations are collected later during the agent’s interaction with the environment.

To begin with, TW-CRL estimates the reward for each state, $ r_{\text{est}}(s_t)$, using the Time-Weighted function (\cref{eq:wt}) and optimizes the reward function  $r_{\phi}$  by minimizing the Contrastive Reward Learning loss (\cref{CRL}). Once the reward function is trained, any suitable reinforcement learning algorithm $\text{POLICY\text{-}OPT}(r, \pi)$ can be used to update the policy $\pi_{\theta}$. We use TD3\citep{TD3} in our experiments. After the initial policy is trained, the agent interacts with the environment and generates new trajectories, which may include both successful and failed demonstrations. These new demonstrations are added to the dataset to further improve the reward function. After updating the reward function, the policy  $\pi_{\theta}$  is retrained based on the updated rewards. This iterative process allows the policy to make better use of the improved reward signal, helping the agent focus on high-reward states while avoiding potential trap states. 

\subsection{Extending TW-CRL to other task types}\label{subsec:extend}
In the discussions above, we assume that the notion of ``goal states'' and ``trap states'' exist in the task. In other words, we assume that the task TW-CRL is solving must be an episodic task with terminal states that clearly signals success or failure. In this section, we discuss how TW-CRL could be applied to tasks beyond this category.

\textbf{Extending to episodic tasks without
clear success / failure terminal states.} For this kind of tasks, we split all trajectories into success and failure according to the total episodic return. Specifically, we apply a threshold $r_\theta$, which is a hyperparameter, to all the expert demonstrations as well as collected trajectories. All trajectories with episodic returns higher than or equal to $r_\theta$ are classified as successful trajectories, while others are classified as failed trajectories.

\textbf{Extending to continuous tasks.} For this kind of tasks, we follow common practice and set a horizon $T$. Each trajectory is then truncated at $T$ timesteps. We can then treat it in the same way as episodic tasks without clear success / failure terminal states as described above. Since the value of $T$ is often defined in the environment (as in all the environments we use introduced in \cref{sec:experiments}), there's no need to tune it as a hyperparameter.

\section{Experiments}\label{sec:experiments}
In this section, we conduct experiments to evaluate the performance of TW-CRL (\cref{subsec:comparison}), visualize and analyze the reward function in TrapMaze-v1 (\cref{subsec:analysis-rewards}), examine TW-CRL's ability to generalize to various goal settings (\cref{gen}), and perform ablation studies (\cref{subsec:ablation}). Details regarding hyperparameters, network architectures, additional training setups, and other information are provided in \cref{ap:envs}, \cref{ap:hyper}, and \cref{ap:license}.

\textbf{Benchmarks}
We use eight environments in total. This includes U-Maze and its two variants TrapMaze-v1 and TrapMaze-v2 which feature unknown trap states and randomized goal locations\cite{fu2020d4rl,gymnasium_robotics2023github},  HumanoidStand, Ant\cite{mujoco},MountainCarCountinuous\cite{gym}, PandaReach, and PandaPush\cite{gallouedec2021pandagym,plappert2018multigoalreinforcementlearningchallenging}. The environments are illustrated in \cref{ap:envs}.

\textbf{Baselines}
We evaluate TW-CRL against six baselines: four IRL methods (GAIL \cite{ho2016generativeadversarialimitationlearning}, AIRL \cite{Fu2017LearningRR}, MaxEnt \cite{Ziebart2008MaximumEI} and MERIT \cite{MERIT}), ICRL \cite{ICRL}, which leverages suboptimal demonstrations, and SASR \cite{ma2024highlyefficientselfadaptivereward}, which incorporates failed demonstrations.

\subsection{Comparison Evaulation}\label{subsec:comparison}
\begin{figure*}[h!]
    \centering
    \begin{subfigure}{0.24\textwidth}
        \centering
        \includegraphics[width=\linewidth]{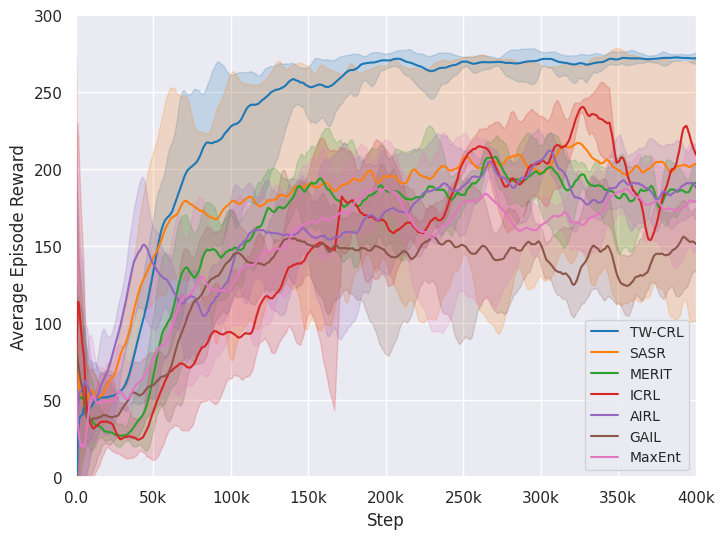}
        \caption{U-Maze}
        \label{fig:traj-map}
    \end{subfigure}
    \begin{subfigure}{0.24\textwidth}
        \centering
        \includegraphics[width=\linewidth]{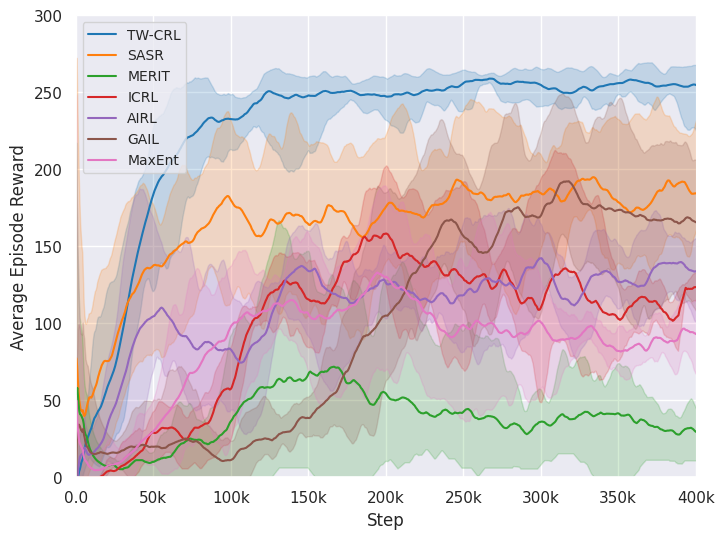}
        \caption{TrapMaze-v1}
        \label{fig:traj-expert}
    \end{subfigure}
    \begin{subfigure}{0.24\textwidth}
        \centering
        \includegraphics[width=\linewidth]{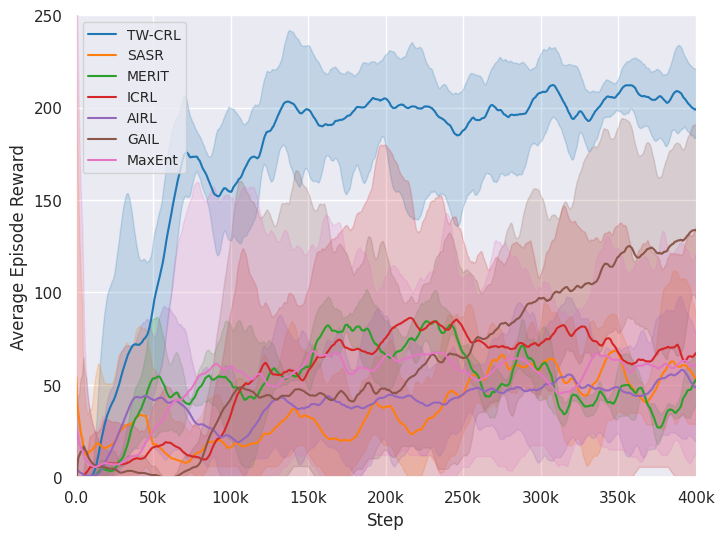}
        \caption{TrapMaze-v1}
        \label{fig:traj-expert}
    \end{subfigure}
    \begin{subfigure}{0.24\textwidth}
        \centering
        \includegraphics[width=\linewidth]{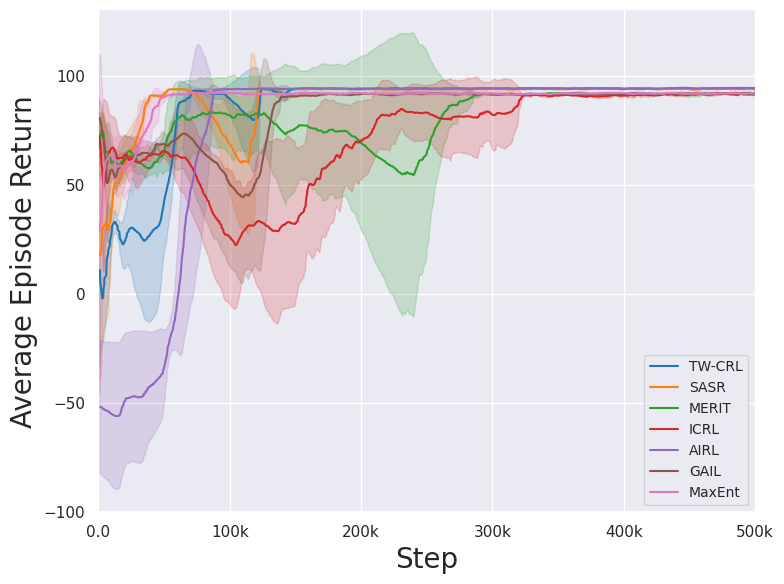}
        \caption{MountainCarCont.}
        \label{fig:traj-mc}
    \end{subfigure}

    \begin{subfigure}{0.24\textwidth}
        \centering
        \includegraphics[width=\linewidth]{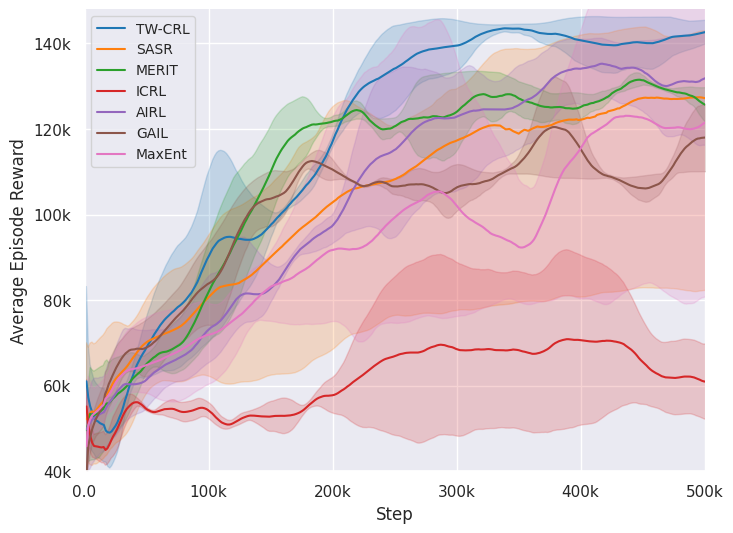}
        \caption{HumanStand}
        \label{fig:traj-shortcut}
    \end{subfigure}
    \begin{subfigure}{0.24\textwidth}
        \centering
        \includegraphics[width=\linewidth]{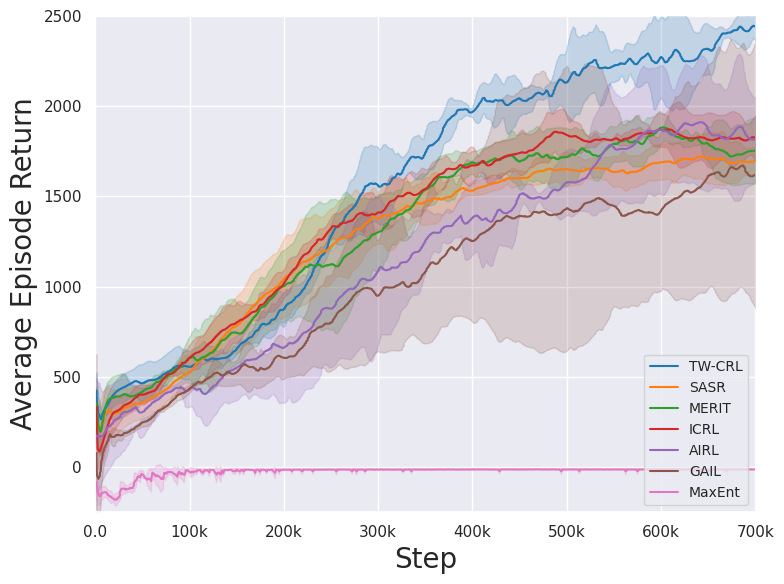}
        \caption{Ant}
        \label{fig:traj-altern}
    \end{subfigure}
    \begin{subfigure}{0.24\textwidth}
        \centering
        \includegraphics[width=\linewidth]{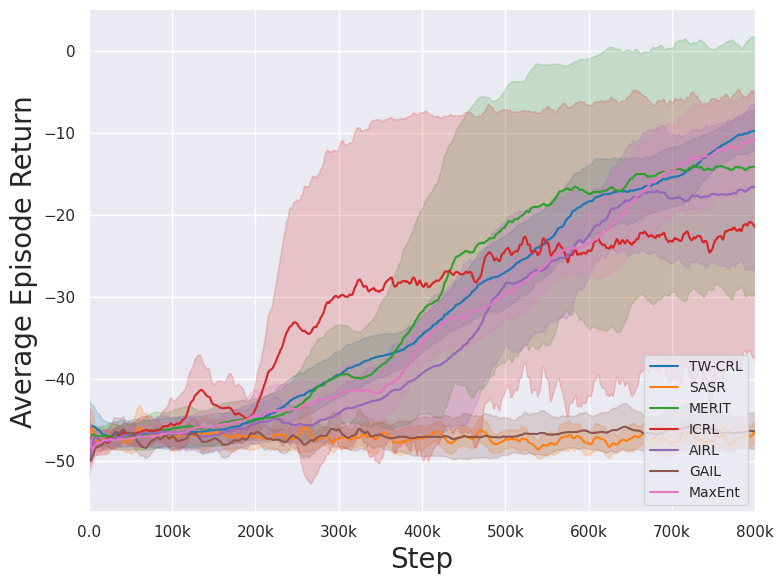}
        \caption{PandaReach}
        \label{fig:traj-expert}
    \end{subfigure}
    \begin{subfigure}{0.24\textwidth}
        \centering
        \includegraphics[width=\linewidth]{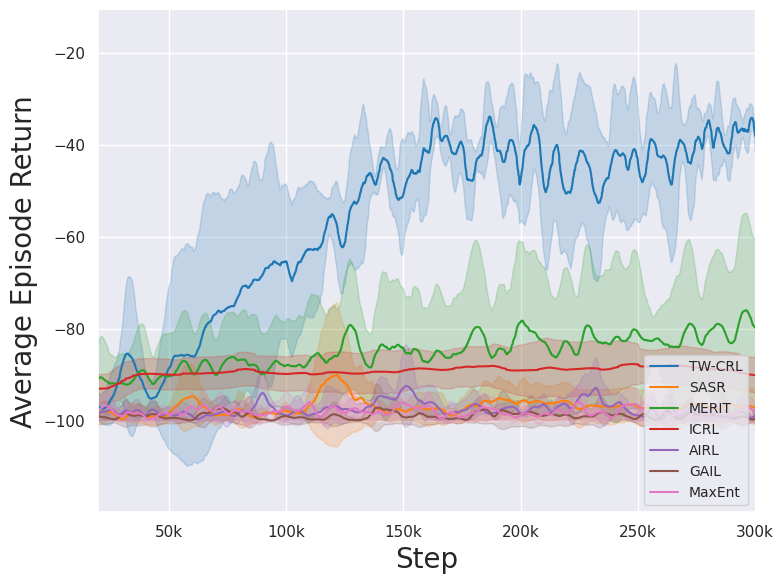}
        \caption{PandaPush}
        \label{fig:traj-expert}
    \end{subfigure}
    \caption{Training curves on benchmarks. The solid curves represent the mean, and the shaded regions indicate the
standard deviation over five runs.}
    \label{fig:main}
\end{figure*}

\cref{fig:main} and \cref{tab:rl_comparison} report the average episodic returns and their variances across all tasks and baselines. We provide five successful demonstrations in the basic U-Maze environment, while TrapMaze-v1 and TrapMaze-v2 each receive 35 successful demonstrations, all ending in a randomly initialized goal located within a single designated grid area of the maze. During training, the same grid area is used for goal initialization, though the specific location within that grid varies each episode. In PandaReach and PandaPush, we offer 10 successful demonstrations. 

\begin{table*}[tbh!]
\caption{Average final return across benchmarks. The best performance is marked in bold, and ± represents the standard deviation over five runs.}
\centering
\small
\begin{tabular}{lcccccccc}
\toprule
\text{Environment}       & \text{U-Maze}                            & \text{TrapMaze-v1}              & \text{TrapMaze-v2}                & \text{MountainCarContinuous}  \\\midrule
\multirow{1}{*}{TW-CRL} 
                         & \textbf{273.05} ± \textbf{1.21}          & \textbf{253.12} ± \textbf{9.46}         & \textbf{201.27} ± \textbf{7.47}            & \textbf{94.23} ± \textbf{0.07}        \\
\multirow{1}{*}{SASR} 
                         & 192.72 ± 78.15          & 196.28 ± 32.89          & 63.68 ± 26.71             & 93.97 ± 0.09        \\\  
\multirow{1}{*}{MERIT} 
                         & 194.72 ± 19.01          & 51.20 ± 43.68          & 68.83 ± 20.59             & 91.70 ± 0.19        \\
\multirow{1}{*}{ICRL}  
                         & 173.73 ± 25.25         & 108.12 ± 36.52          & 72.60 ± 26.35             & 92.02 ± 0.10        \\
\multirow{1}{*}{GAIL} 
                         & 138.64 ± 20.25          & 159.06 ± 25.53          & 113.94 ± 50.26             & 91.80 ± 0.39        \\ 
\multirow{1}{*}{AIRL} 
                        & 208.21 ± 17.83          & 160.22 ± 11.25          & 48.95 ± 19.92             & 93.86 ± 0.53        \\
\multirow{1}{*}{MaxEnt} 
                        & 161.28 ± 41.38          & 129.70 ± 18.50          & 77.89 ± 30.10             & 91.71 ± 0.17        \\ \midrule

\text{Environment}       & \text{HumanStand ($\times10^3$)}                            & \text{Ant ($\times10^2$)}              & \text{PandaReach}                & \text{PandaPush}  \\\midrule
\multirow{1}{*}{TW-CRL} 
                         & \textbf{144.69} ± \textbf{1.05}          & \textbf{24.26} ± \textbf{0.12}         & \textbf{-5.98} ± \textbf{0.55}            & \textbf{-34.89} ± \textbf{6.58}        \\
\multirow{1}{*}{SASR} 
                         & 128.22 ± 22.9         & 17.62 ± 0.87          & -47.41 ± 0.51             & -96.77 ± 2.34        \\
\multirow{1}{*}{MERIT} 
                         & 135.39 ± 3.37          & 17.34 ± 1.15          & -11.35 ± 19.85             & -76.09 ± 14.64        \\
\multirow{1}{*}{ICRL}  
                         & 57.89 ± 7.67          & 18.14 ± 0.40          & -24.00 ± 12.83             & -88.32 ± 3.42        \\
\multirow{1}{*}{GAIL} 
                         & 121.32 ± 9.35          & 15.22 ± 3.98          & -46.53 ± 2.28             & -99.00 ± 1.40        \\ 
\multirow{1}{*}{AIRL} 
                        & 134.83 ± 10.45          & 18.72 ± 1.43          & -10.76 ± 5.93             & -97.39 ± 3.00        \\
\multirow{1}{*}{MaxEnt} 
                        & 116.63 ± 20.82         & -0.12 ± 0.01          & -5.97 ± 2.28             & -96.45 ± 2.54        \\

\bottomrule
\end{tabular}
\label{tab:rl_comparison}
\end{table*}

\begin{figure*}[tbh!]
    \centering
    \begin{subfigure}{0.19\textwidth}
        \centering
        \includegraphics[width=\linewidth]{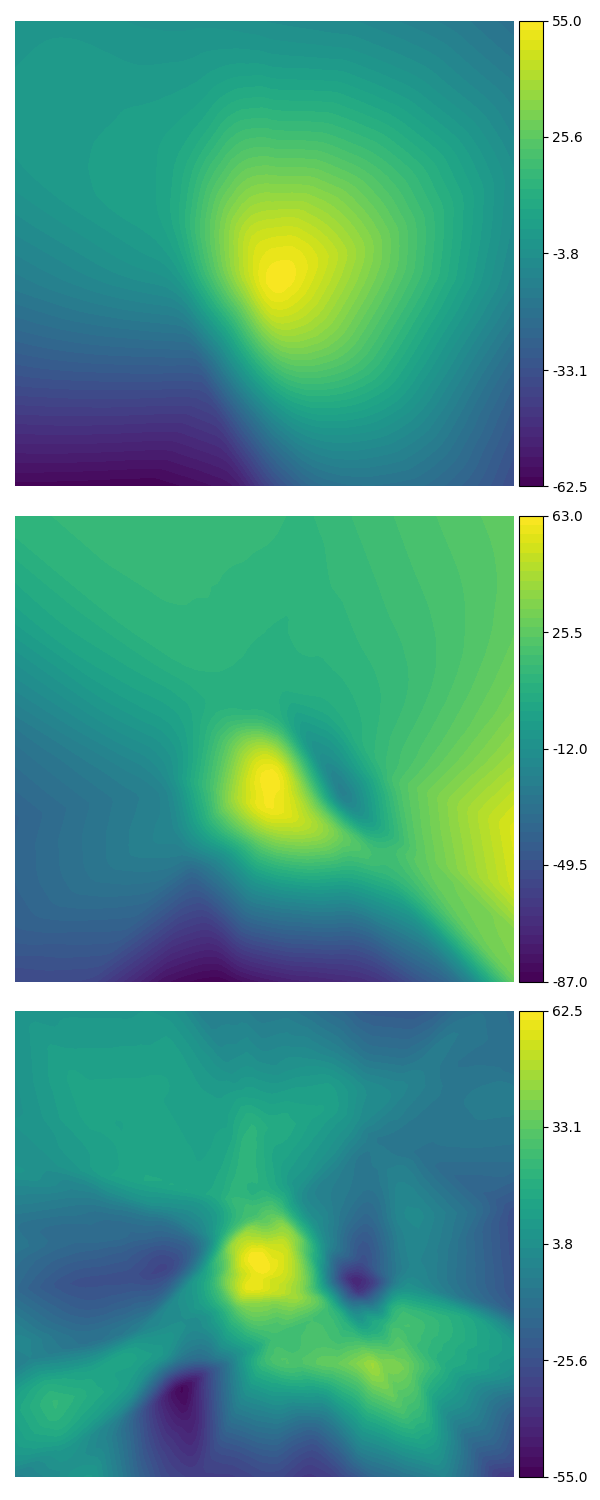}
        \caption{TW-CRL}
        \label{fig:TW-CRL}
    \end{subfigure}
    \hfill
    \begin{subfigure}{0.19\textwidth}
        \centering
        \includegraphics[width=\linewidth]{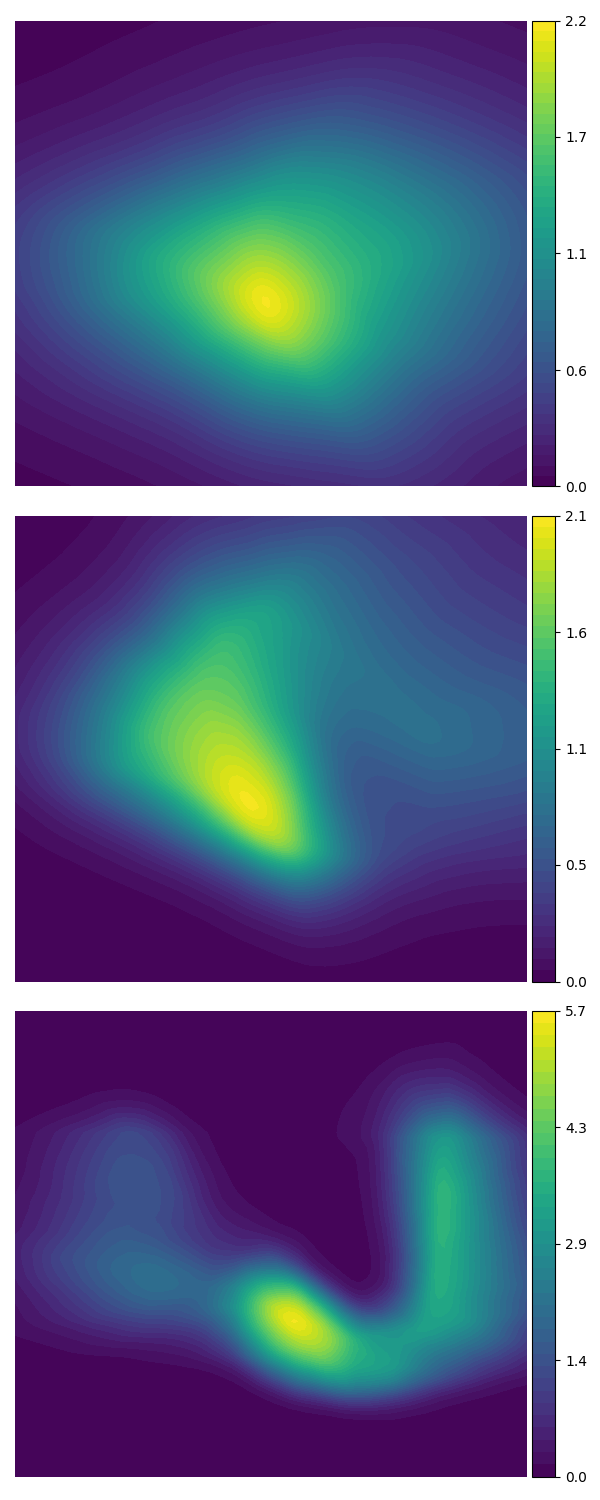}
        \caption{GAIL}
        \label{fig:GAIL}
    \end{subfigure}
    \hfill
    \begin{subfigure}{0.19\textwidth}
        \centering
        \includegraphics[width=\linewidth]{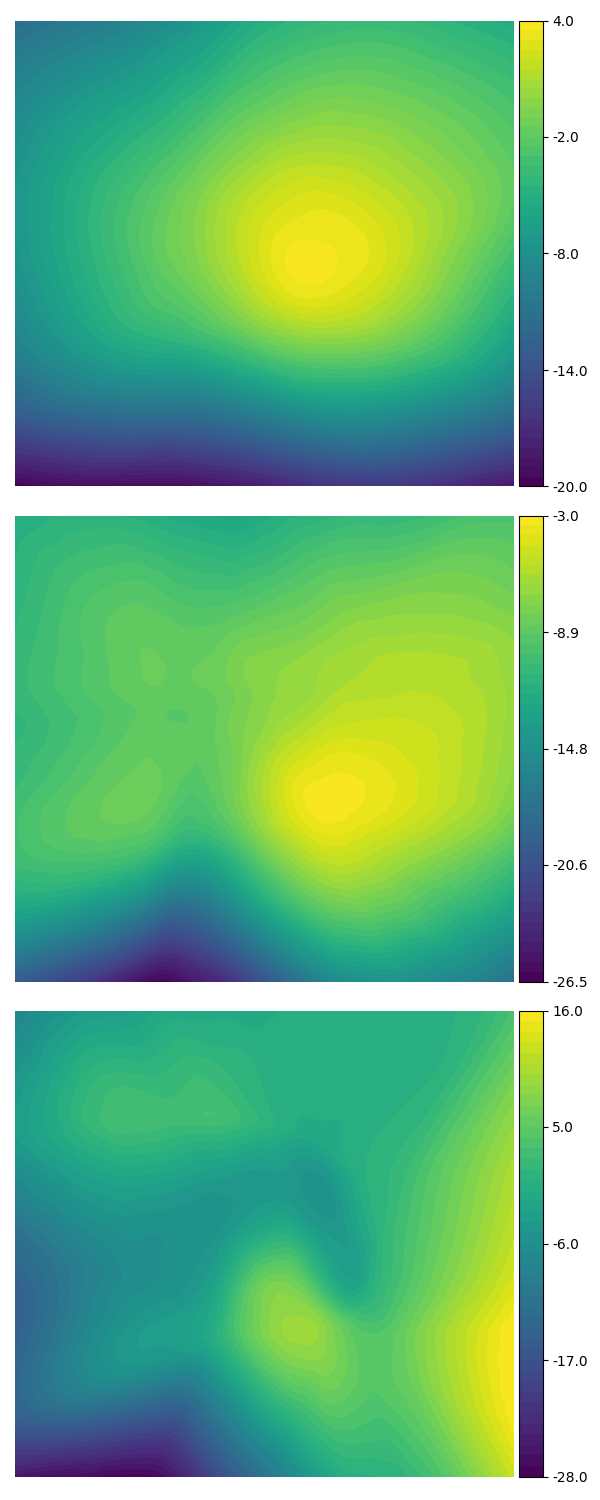}
        \caption{AIRL}
        \label{fig:AIRL}
    \end{subfigure}
    \hfill
    \begin{subfigure}{0.19\textwidth}
        \centering
        \includegraphics[width=\linewidth]{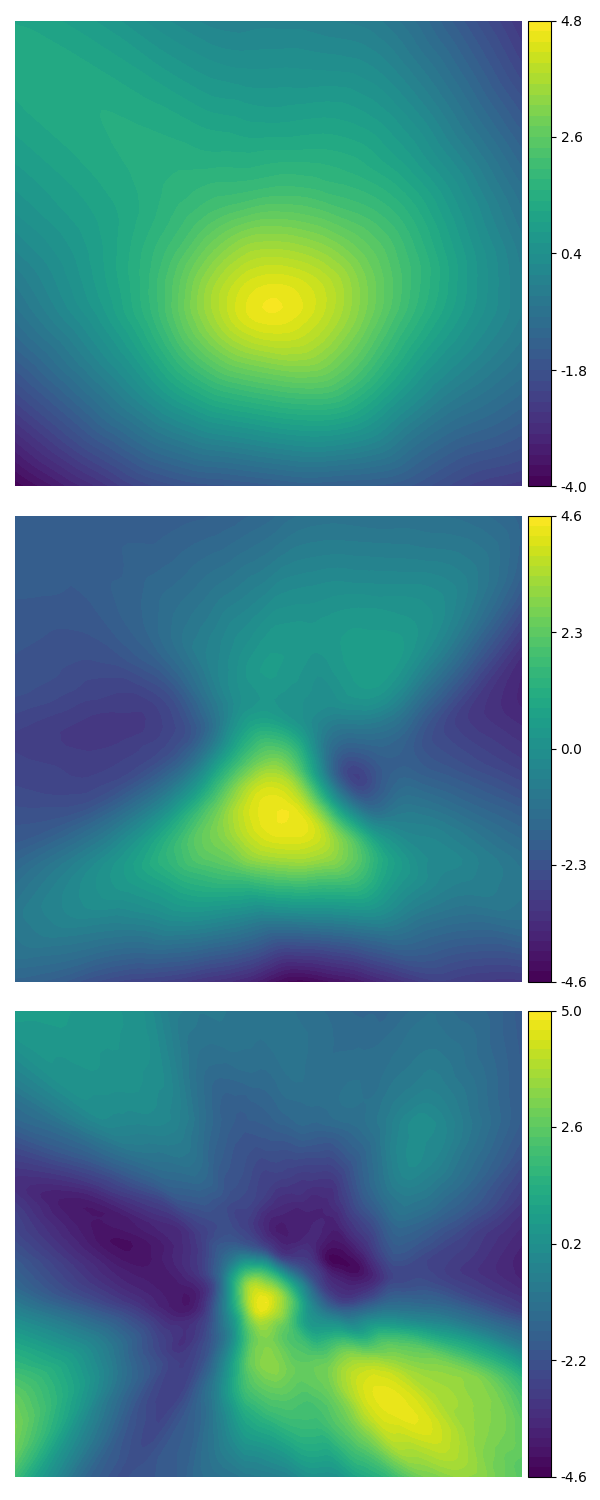}
        \caption{MaxEnt}
        \label{fig:MaxEnt}
    \end{subfigure}
    \hfill
    \begin{subfigure}{0.19\textwidth}
        \centering
        \includegraphics[width=\linewidth]{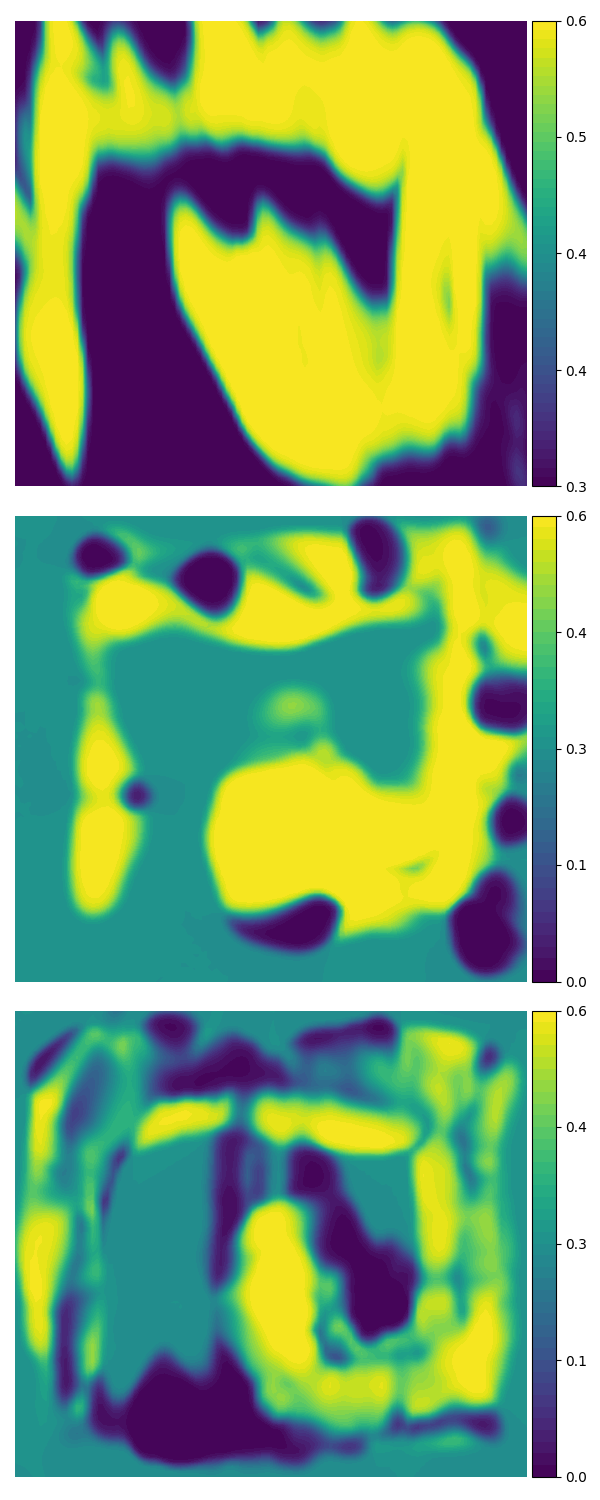}
        \caption{SASR}
        \label{fig:SASR}
    \end{subfigure}
    \caption{Visualization of the reward function in the TrapMaze-v1 environment for TW-CRL and baseline methods. Each column represents a different method, and each row shows a training stage, with the final row illustrating the fully trained reward functions.}
    \label{fig:mid_rew}
\end{figure*}

The results indicate that the performance of TW-CRL is better than or comparable to all baselines in terms of average episode return, with faster convergence and reduced variance. For example, in the U-Maze environment, the average final return in our algorithm demonstrates enhancements of $41.71\%$, $40.22\%$, $57.16\%$, $96.96\%$, $31.15\%$, and $69.33\%$ over SASR, MERIT, ICRL, GAIL, AIRL, and MaxEnt, respectively.

\subsection{Analysis of Learned Rewards}\label{subsec:analysis-rewards}
In this section, we analyze the learned reward functions across different methods using the TrapMaze-v1 environment to illustrate why TW-CRL outperforms other approaches. The simplified illustration of the TrapMaze-v1 environment map is shown in \cref{fig:traj-map}, where the red area represents the goal states, the blue area represents the trap states, and the gray blocks represent walls that the agent cannot get through. We provide a total of 35 successful demonstrations in which agents follow the main path to the goal, as illustrated in \cref{fig:traj-expert}. \cref{fig:mid_rew} visualizes the reward functions learned by different methods throughout the training process. An exhaustive figure containing all baseline methods is shown in Appendix~\ref{ap:ad-reward}.

The reward function visualizations reveal that GAIL, AIRL, and MaxEnt fail to recognize unseen trap states, and give high rewards only to states seen in successful demonstrations. As a result, these methods limit the agent to only imitating expert behavior without discovering better ways to reach the goal. On the other hand, SASR can detect unseen trap states by using failed demonstrations. It primarily assigns higher rewards to states that frequently appear in successful trajectories and lower rewards to those that frequently appear in failed trajectories. However, this approach still limits the agent’s ability to explore and find better solutions.

TW‑CRL overcomes these limitations by enabling the agent to avoid trap states and efficiently discover paths to the goal. This is achieved by 1) utilizing both successful and failed trajectories in the Contrastive Reward Learning process and 2) utilizing temporal information in our Time-Weighted function to only place high emphasis on later states that may have heavier impacts on the outcome of the trajectory. As a result, TW‑CRL not only identifies both goal and trap states more accurately than imitation‑only methods, but also encourages broader exploration—allowing the agent to uncover more efficient routes instead of confining themselves to expert demonstrations. For example, the shortcut in \cref{fig:traj-shortcut} and the alternative paths in \cref{fig:traj-altern} were never shown in the expert demonstrations, but TW-CRL allows the agent to explore these paths. 
\begin{figure}[h!]
    \centering
    \begin{subfigure}{0.21\textwidth}
        \centering
        \includegraphics[width=\linewidth]{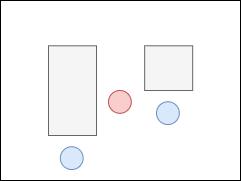}
        \caption{TrapMaze-v1 map}
        \label{fig:traj-map}
    \end{subfigure}
    \hfill
    \begin{subfigure}{0.21\textwidth}
        \centering
        \includegraphics[width=\linewidth]{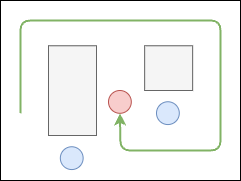}
        \caption{An expert demo}
        \label{fig:traj-expert}
    \end{subfigure}
    \hfill
    \begin{subfigure}{0.21\textwidth}
        \centering
        \includegraphics[width=\linewidth]{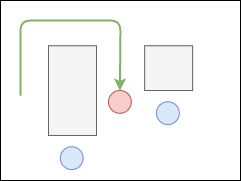}
        \caption{A shortcut path}
        \label{fig:traj-shortcut}
    \end{subfigure}
    \hfill
    \begin{subfigure}{0.21\textwidth}
        \centering
        \includegraphics[width=\linewidth]{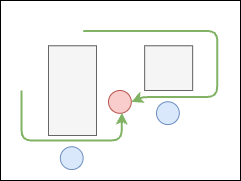}
        \caption{Alternative paths}
        \label{fig:traj-altern}
    \end{subfigure}
    \caption{Illustration of the map and potential trajectories in TrapMaze-v1.}
    \label{fig:traj}
\end{figure}

\begin{table}[h!]
\centering
\caption{Average return under three goal reset settings. 1, 3, and any. \#goal represent for the number of grids where the goal state might be in.
}
\footnotesize
\begin{tabular}{cccccc}
\toprule
\text{\# goal state(s)}    & \text{TW-CRL}                           & \text{SASR}             & \text{MERIT}             & \text{ICRL}                & \text{GAIL}                          \\ \midrule
\multirow{1}{*}{1}
                         & \textbf{253.12} ± \textbf{9.46}          & 196.28 ± 32.89          & 51.20 ± 50.95      & 108.12 ± 36.52           & 159.06 ± 25.53 \\  
\multirow{1}{*}{3} 
                         & \textbf{240.48 } ± \textbf{4.44}          & 150.84 ± 50.95           & 96.37 ± 11.94      & 39.59 ± 17.93          & 59.79 ± 22.87 \\
\multirow{1}{*}{Any} 
                         & \textbf{221.33} ± \textbf{11.58}         & 122.78 ± 20.39           & 67.93 ± 23.7     & 90.12 ± 22.56            & 55.25 ± 21.67 \\

\bottomrule
\end{tabular}
\label{tab:multi_goal}
\end{table}

\subsection{Generalization Analysis} \label{gen}
We then evaluate TW-CRL’s ability to generalize beyond the goal configurations provided in the original demonstrations in the TrapMaze-v1 environment, where the goal is randomly placed within one or several grid cells. Specifically, we provide demonstrations where the goal is initialized within a single grid cell, but during training and testing, the goal is allowed to be initialized within 1, 3, or any grid cells in the maze. \cref{tab:multi_goal} reports the average returns for each configuration. While all methods experience some drop in performance as the distribution of possible goal states broadens, TW-CRL remains consistently superior to the baselines, suggesting that its learned reward function can adapt successfully to unseen goal locations. Further results and details can be found in \cref{ap:envs} and \cref{ap:ad-generalization}.

\subsection{Ablation Study}\label{subsec:ablation}
\begin{figure}[h!]
    \centering
    \begin{subfigure}{0.49\textwidth}
        \centering
        \includegraphics[width=0.49\linewidth]{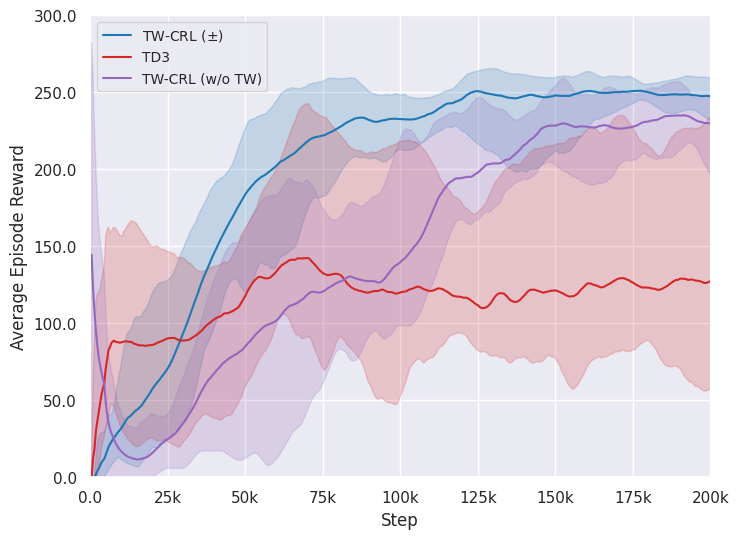}
        \includegraphics[width=0.49\linewidth]{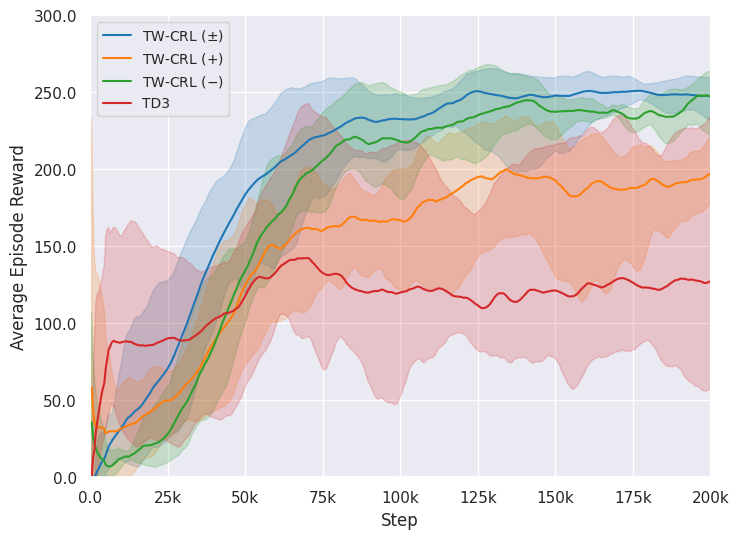}
        \caption{TrapMaze-v1}
        \label{fig:ablation-v1}
    \end{subfigure}
    \hfill
    \begin{subfigure}{0.49\textwidth}
        \centering
        \includegraphics[width=0.49\linewidth]{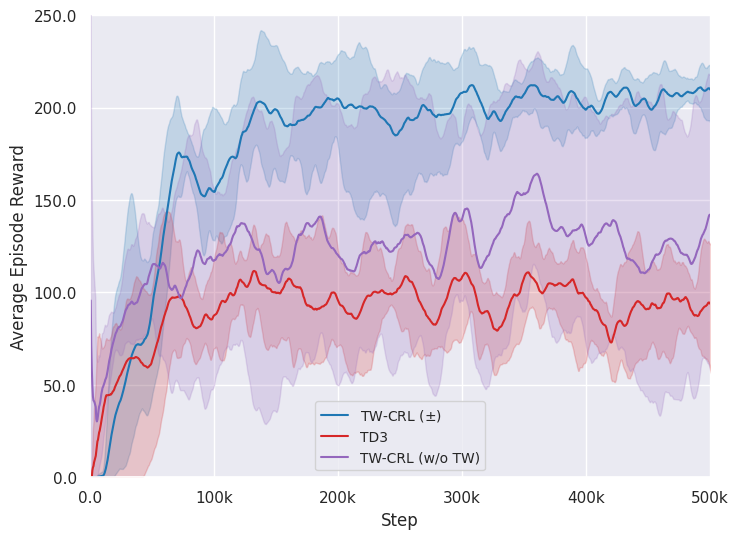}
        \includegraphics[width=0.49\linewidth]{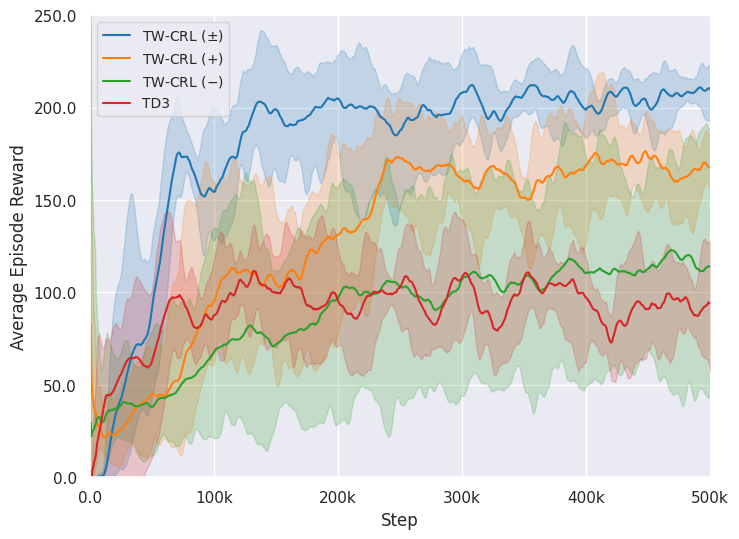}
        \caption{TrapMaze-v2}
        \label{fig:ablation-v2}
    \end{subfigure}

    \caption{Ablation studies in TrapMaze-v1 and TrapMaze-v2 environments. The left figures show the ablation of the Time-Weighted function, the right figures show the ablation of the Contrastive Reward Learning loss function.}
    \label{fig:ablation}
\end{figure}

\textbf{Effect of the Time-Weighted function}
\cref{fig:ablation-v1} and \cref{fig:ablation-v2} (left) compare TW-CRL with the Time-Weighted function (blue) to a simpler approach using constant rewards of $+1$ for success states and $-1$ for failure states (purple). Results show that using time-weighted rewards accelerates convergence and achieves higher final returns in both TrapMaze-v1 and TrapMaze-v2. The advantage is even clearer in TrapMaze-v2, where the time-weighted approach significantly outperforms the non-time-weighted version and vanilla TD3. These findings suggest that time weighting is particularly beneficial in complex tasks.

\textbf{Effect of using both successful and failed demonstrations}
\cref{fig:ablation-v1} and \cref{fig:ablation-v2} (right) compare four different settings: (1) TW-CRL with both successful and failed demonstrations (blue), (2) TW-CRL with only successful demonstrations (orange), (3) TW-CRL with only failed demonstrations (green), and (4) vanilla TD3 without demonstrations (red). The results show that using both successful and failed demonstrations (blue) achieves the fastest convergence and highest final reward. In contrast, using only successful (orange) or failed (green) demonstrations results in slower learning and slightly lower performance, though both still surpass TD3 (red). This trend is more pronounced in the more complex TrapMaze-v2 environment, confirming that combining both types of demonstrations provides a stronger learning signal.
\section{Conclusion}
In this paper, we propose TW-CRL, an inverse reinforcement learning approach that infers reward functions by leveraging temporal information from both successful and failed demonstrations. By incorporating the Time-Weighted function and Contrastive Reward Learning, TW-CRL learns an accurate and informative reward function, enabling the agent to efficiently locate goals while avoiding repeated failures, thereby enhancing learning efficiency and increasing task success rates. Additionally, TW-CRL enhances exploration beyond expert demonstrations by discovering alternative high-value paths or shortcuts that were not explicitly demonstrated. Empirical evaluations demonstrate that TW-CRL outperforms baseline methods, showcasing its effectiveness in enhancing agent performance and learning efficiency in complex tasks.

\bibliography{neurips_2025}

\newpage
\appendix
\onecolumn

\section{Mathematical Derivations}
\subsection{Derivation of \cref{eq:ftT}}\label{ap:der-1}
Consider a trajectory $\tau = \{(s_0, a_0), (s_1, a_1), ..., (s_{T-1}, a_{T-1})\}$ with horizon $T$. Recall from \cref{subsec:trapstates} that:
\begin{align}
    \Sgtrap = \{\sgtrap \in {\mathcal{S}-\mathcal{S}_{\text{goal}}} \mid \sum\limits_{s' \in \Sgtrap} P( s' \mid \sgtrap, a) = 1, \forall a \in \Action \}.
\end{align}
where ${\mathcal{S}}$ is the state space, $\mathcal{S}_{\text{goal}}$ is the set of all goal states, $\Sgtrap\subseteq{\mathcal{S}}$ is the set of all trap states. We assume the following:
\begin{enumerate}
    \item The initial state $s_0\notin\Sgtrap$. 
    \item For each timestep $t=1,2,...,T-1$, $P(s_{t}\in\Sgtrap|s_{t-1}\notin\Sgtrap) = k$. For now, we assume that $k\in\mathbb{R}$ is a constant value.
\end{enumerate}
Given these assumptions, we can model the dynamics of the agent as an absorbing Markov chain, where the trap states are absorbing states and non-trap states are transient states. We can derive the probability of the agent being in a trap state at timestep $t$:
\begin{align}
    P(s_t\in\Sgtrap) &= 1 - P(s_t\notin\Sgtrap) \\
    &= 1 - P(s_{t-1}\notin\Sgtrap)P(s_t\notin\Sgtrap|s_{t-1}\notin\Sgtrap) \\
    &= 1-P(s_0\notin\Sgtrap)\prod\limits_{i=1}^{t}P(s_i\notin\Sgtrap|s_{i-1}\notin\Sgtrap) \\
    &= 1-\prod\limits_{i=1}^{t}P(s_i\notin\Sgtrap|s_{i-1}\notin\Sgtrap) \\
    &= 1-\prod\limits_{i=1}^{t}[1-P(s_i\in\Sgtrap|s_{i-1}\notin\Sgtrap)] \\
    &= 1-(1-k)^t
\end{align}

In a failed demonstration, recall from \cref{subsec:trapstates} that we assume the final state is always a trap state ($s_T\in\Sgtrap$). Therefore, given the prior knowledge that a demonstration is a failed demonstration, we can derive the conditional probability of a state $s_t$ of timestep $t$ being a trap state as follows:
\begin{align}
    P(s_t\in\Sgtrap | s_T\in\Sgtrap) &= \frac{P(s_t\in\Sgtrap, s_T\in\Sgtrap)}{P(s_T\in\Sgtrap)} \\
    &= \frac{P(s_t\in\Sgtrap)}{P(s_T\in\Sgtrap)} \\
    &= \frac{1-(1-k)^t}{1-(1-k)^T}
\end{align}

We can extend the conclusion above to cases where $k$ is decided by the current timestep $t$ and horizon $T$. Formally,
\begin{align}
    P(s_t\in\Sgtrap | s_T\in\Sgtrap) &= \frac{1-(1-f(t, T))^t}{1-(1-f(t, T))^T}
\end{align}

\subsection{Derivation of \cref{eq:wt-f} and (\ref{eq:wt-s})}\label{ap:der-23}

We first show how \cref{eq:wt-f} is derived. By applying \cref{eq:ftT} to \cref{eq:pst} we have:
\begin{align}
    P(s_t\in\Sgtrap | s_T\in\Sgtrap) &= \frac{1-(1-\frac{e^{\alpha t} -1 }{e^{\alpha T} -1})^t}{1-(1-\frac{e^{\alpha t} -1 }{e^{\alpha T} -1})^T}
\end{align}

To simplify computation, notice that $1-(1-\frac{e^{\alpha t} -1 }{e^{\alpha T} -1})^T \approx 1$ when $T$ is large. Therefore,

\begin{align}
P(s_t\in\Sgtrap | s_T\in\Sgtrap) &\approx 1-(1-\frac{e^{\alpha t} -1 }{e^{\alpha T} -1})^t
\end{align}

Similarly, to derive \cref{eq:wt-s}, we apply the same process for $P(s_t\in\Sgoal | s_T\in\Sgoal)$:
\begin{align}
    P(s_t\in\Sgoal | s_T\in\Sgoal) &= \frac{1-(1-\frac{e^{\alpha t} -1 }{e^{\alpha T} -1})^t}{1-(1-\frac{e^{\alpha t} -1 } {e^{\alpha T} -1})^T} \\
    &\approx 1-(1-\frac{e^{\alpha t} -1 }{e^{\alpha T} -1})^t
\end{align}

\subsection{Plotting the Time-Weighted function}\label{ap:wt}

\begin{figure}[h!]
    \centering
    \includegraphics[width=0.5\linewidth]{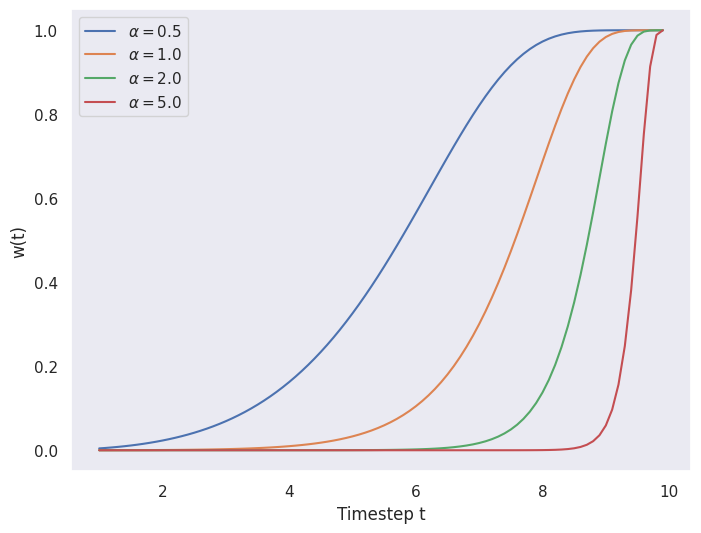}
    \caption{Illustration of $w(t)$ with T=10 and various $\alpha$ values.}
    \label{fig:wt}
\end{figure}

Recall that in our experiments, we use the following function as our Time-Weighted function:
\begin{align}\label{eq:wt-ap}
    w(t) = 1-(1-\frac{e^{\alpha t} -1 }{e^{\alpha T} -1})^t
\end{align}
A plot of the above function with different values of $\alpha$ is shown in \cref{fig:wt}. We can observe that as the value of $\alpha$ increases, the curve becomes steeper and moves towards later timesteps, indicating that the Time-Weighted function $w(t)$ places a higher significance on later timesteps. In practice, the value of $\alpha$ can be tuned as a hyperparameter. 

\subsection{Justification for \cref{eqn:assign}}\label{ap:crl}

Consider the derivative of the Contrastive Reward Learning loss function (\cref{CRL}):
\begin{align}
\frac{d}{d \phi} \mathcal{L}_{\text{CRL}} 
&= \frac{d \mathcal{L}_{\text{CRL}}}{d r_{\phi}(s_t)} \frac{d r_{\phi}(s_t)}{d \phi} \\
&\propto 2N_{\text{goal}}(r_{\phi}(s_t) - \Pgoal) \notag \\
&\quad + 2N_{\text{trap}}(r_{\phi}(s_t) + \Ptrap)). \label{eqn:idea_loss}
\end{align}

By setting the derivative in \cref{eqn:idea_loss} to zero, we obtain:
\begin{equation} \label{eqn:derivative}
r_{\phi}(s_t) = \frac{N_{\text{goal}} \Pgoal - N_{\text{trap}} \Ptrap}{N_{\text{goal}} + N_{\text{trap}}}.
\end{equation}
If the number of successful and failed samples is equal, i.e., $N_{\text{goal}} = N_{\text{trap}}$ , equation \cref{eqn:derivative} simplifies to:

\begin{equation}
    r_{\phi}(s_t) = \frac{\Pgoal - \Ptrap}{2}.
\end{equation}

This result shows that the learned reward function  $r_{\phi}$  is proportional to the ideal target  $\Pgoal - \Ptrap$ . Therefore, the learned function effectively captures the difference between the success and failure probabilities.

If the number of successful and failed samples is not equal, i.e.,  $N_{\text{goal}} \neq N_{\text{trap}}$, the reward function still approximates  $\Pgoal - \Ptrap$, with a scaling factor determined by the ratio of successful to failed samples. This is because  $r_{\phi}$  is optimized as a weighted difference between the probabilities of success and failure.

Therefore, by training with the labeling method in \cref{eqn:assign},  we can achieve an equivalent effect to \cref{eqn:ideal}, resulting in a reward function that approximates the desired goal-trap probability difference.

\section{Pseudocode}

\begin{algorithm*}[h!]                    
  \caption{TW-CRL}
  \label{alg:tw-crl}
  \begin{algorithmic}[1]               
    \State \textbf{Input:} initial reward $r_{\phi}$, policy $\pi_{\theta}$,
           updater $\textsc{PolicyOpt}(r,\pi)$, expert demos $\mathcal{D}_\mathrm{e}$
    \State $R_{\phi} \gets [\,]$, \ $R_{\text{est}} \gets [\,]$  \Comment{Initialize empty lists}
    \State $\mathcal{D} \gets \mathcal{D}_\mathrm{e}$
    \While{convergence not achieved}
      \For{each state $s_t \in \mathcal{D}$}
      \Comment{Estimate reward label for each state}
        \State $w(t) =  1-(1-\frac{e^{\alpha t} -1 }{e^{\alpha T} -1})^t$
        \If{$s_t$ is in successful trajectories}
          \State $r_{\text{est}}(s_t) = \phantom{-}w(t)$
        \Else
          \State $r_{\text{est}}(s_t) = -\,w(t)$
        \EndIf
        \State $R_{\text{est}}\!\gets R_{\text{est}}\cup\{r_{\text{est}}(s_t)\}$,
               \ $R_{\phi}\!\gets R_{\phi}\cup\{r_{\phi}(s_t)\}$ \Comment{Append $r_{\text{est}}(s_t)$, $r_{\phi}(s_t)$ to $R_\text{est}$, $R_{\phi}$}
      \EndFor
      \State $\mathcal{L}_{\text{CRL}}
        = \dfrac{1}{N}\sum_{i=1}^{N}\!\bigl(R_{\phi}[i]-R_{\text{est}}[i]\bigr)^{2}$ \Comment{Update reward function $r_{\phi}$ with $\mathcal{L}_{\text{CRL}}$} 
      \State Update $r_{\phi}$ by minimising $\mathcal{L}_{\text{CRL}}$
      \State $\theta \gets \textsc{PolicyOpt}(r_{\phi},\pi_{\theta})$ \Comment{Update policy $\pi_{\theta}$}
      \State $\mathcal{D} \gets \mathcal{D}\cup\mathcal{D}^{+}\cup\mathcal{D}^{-}$ \Comment{Collect new successful and failed demos $\mathcal{D}^+$, $\mathcal{D}^-$}
    \EndWhile
  \end{algorithmic}
\end{algorithm*}

\section{Experiment} 
\subsection{Experiment Environments} \label{ap:envs}

\begin{figure*}[h]
    \centering
    \begin{subfigure}{0.24\textwidth}
        \centering
        \includegraphics[width=\linewidth]{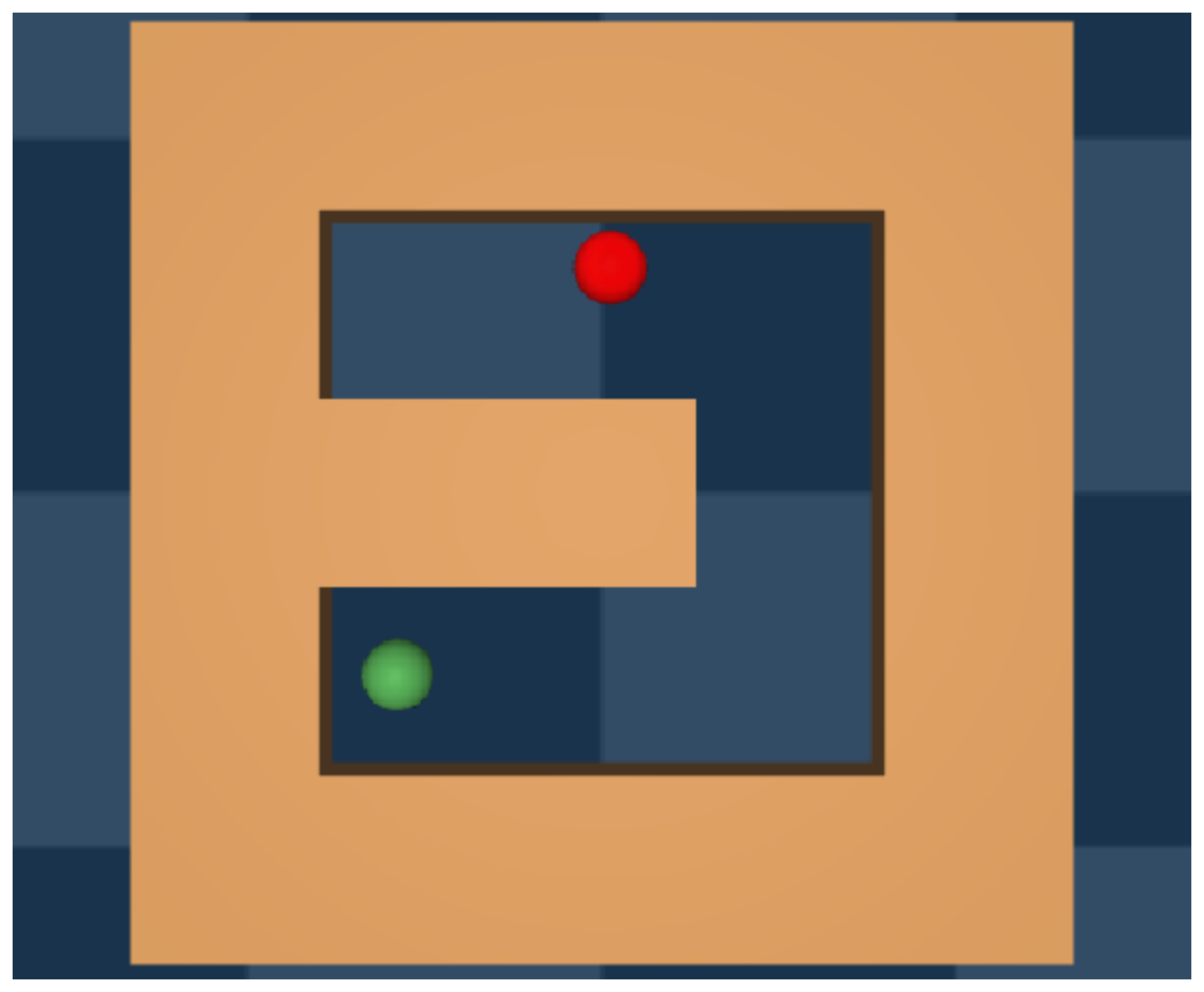}
        \caption{U-Maze}
    \end{subfigure}
    \begin{subfigure}{0.24\textwidth}
        \centering
        \includegraphics[width=\linewidth]{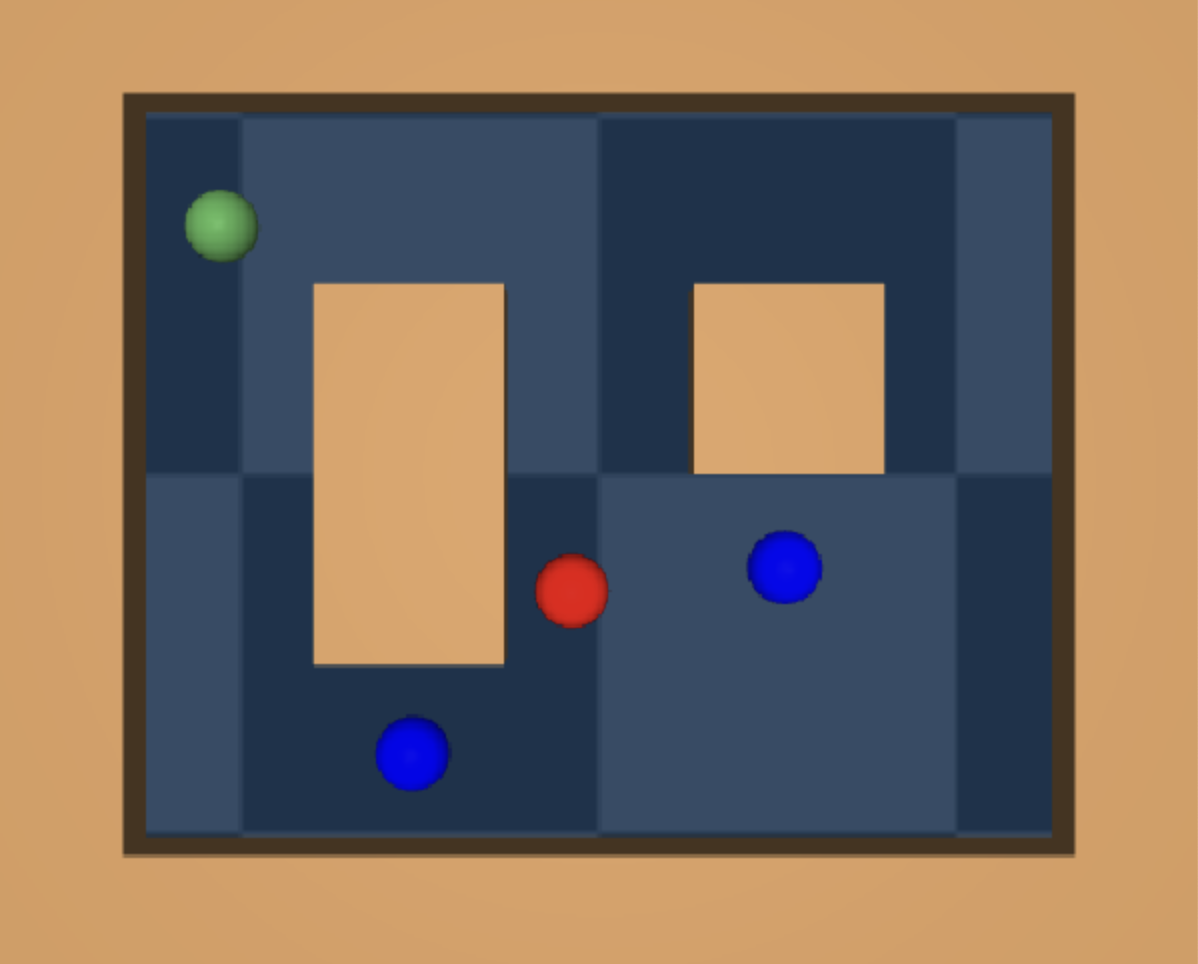}
        \caption{TrapMaze-v1}
    \end{subfigure}
    \begin{subfigure}{0.24\textwidth}
        \centering
        \includegraphics[width=\linewidth]{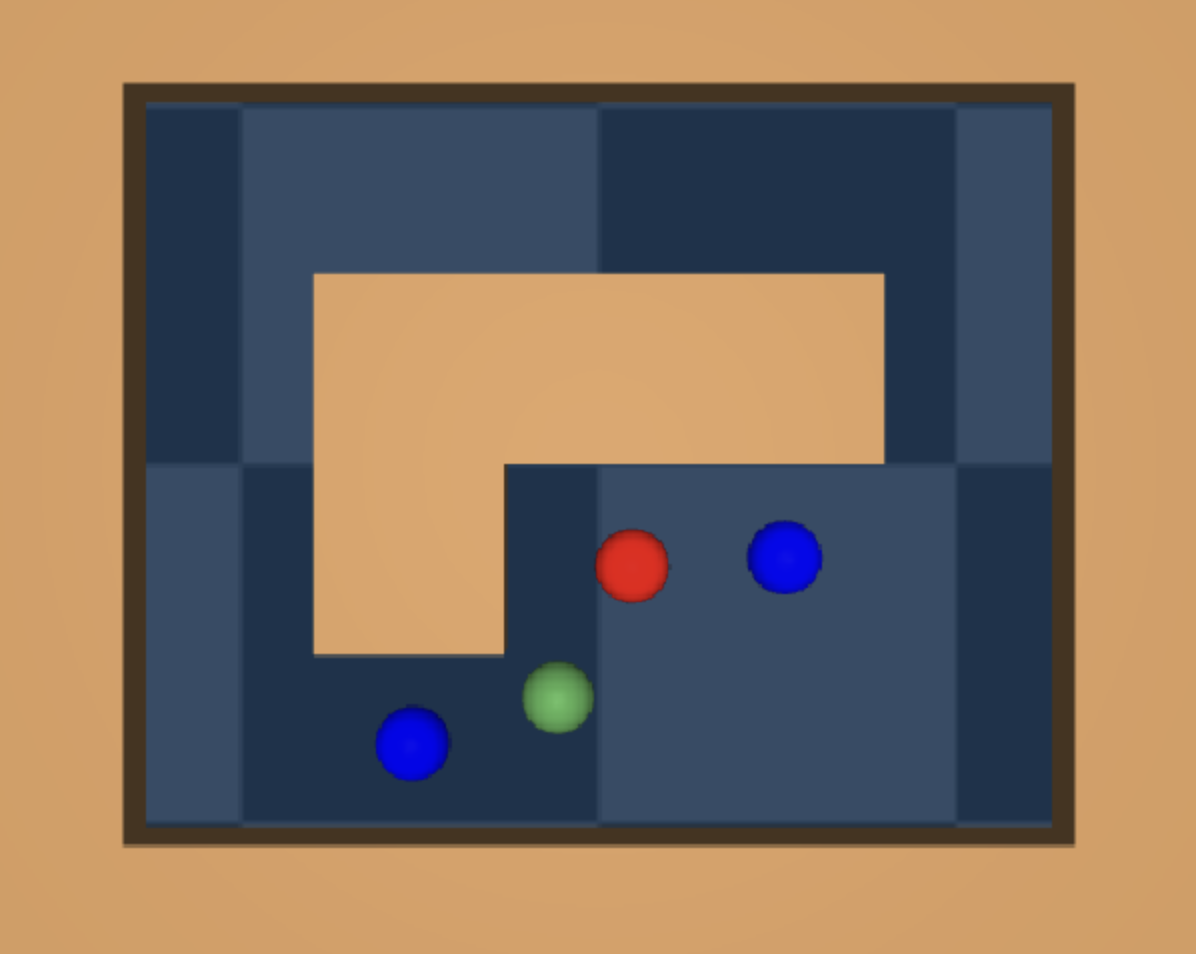}
        \caption{TrapMaze-v2}
    \end{subfigure}
    \begin{subfigure}{0.24\textwidth}
        \centering
        \includegraphics[width=\linewidth]{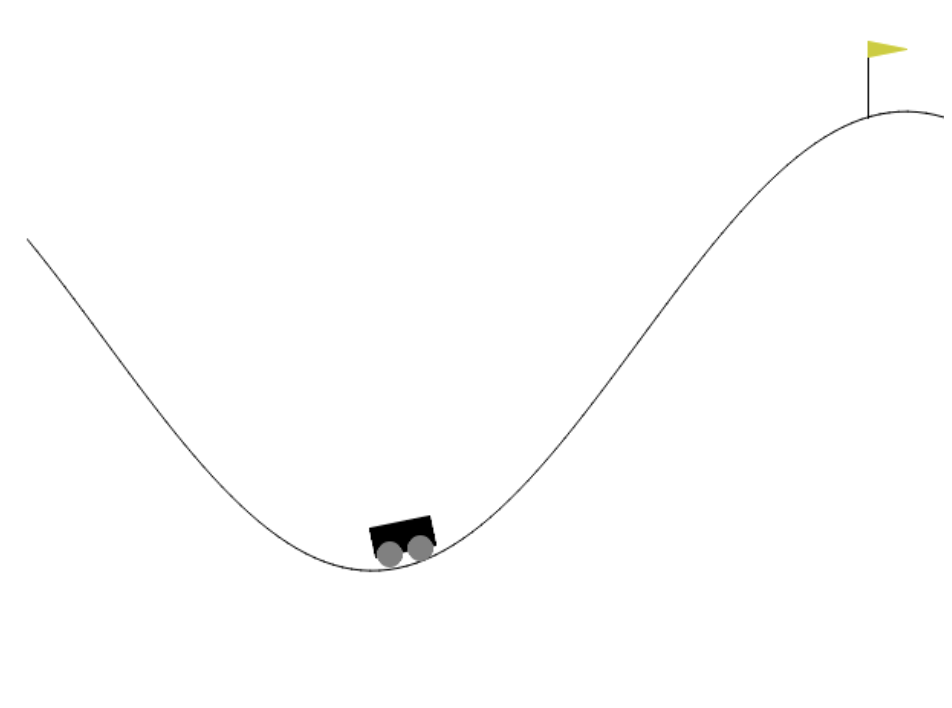}
        \caption{MountainCar-v0}
    \end{subfigure}

    \begin{subfigure}{0.24\textwidth}
        \centering
        \includegraphics[width=\linewidth]{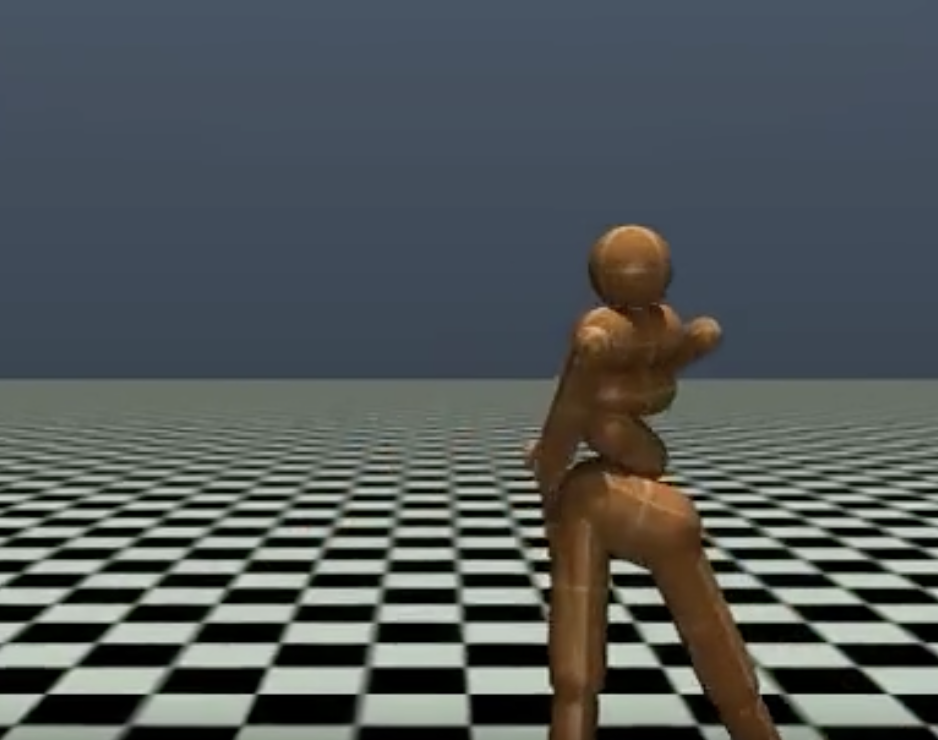}
        \caption{HumanStand-v5}
    \end{subfigure}
    \begin{subfigure}{0.24\textwidth}
        \centering
        \includegraphics[width=\linewidth]{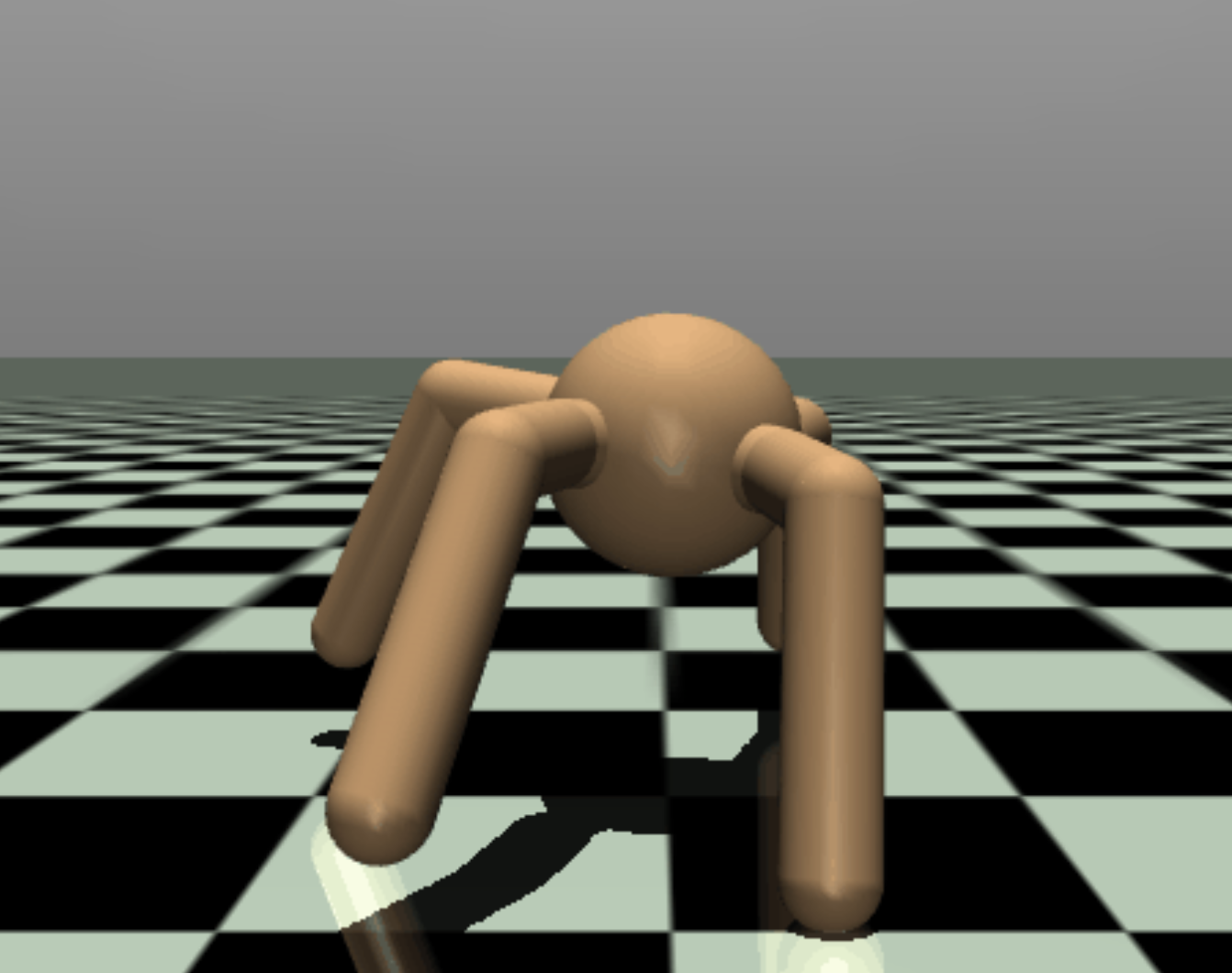}
        \caption{Ant-v5}
    \end{subfigure}
    \begin{subfigure}{0.24\textwidth}
        \centering
        \includegraphics[width=\linewidth]{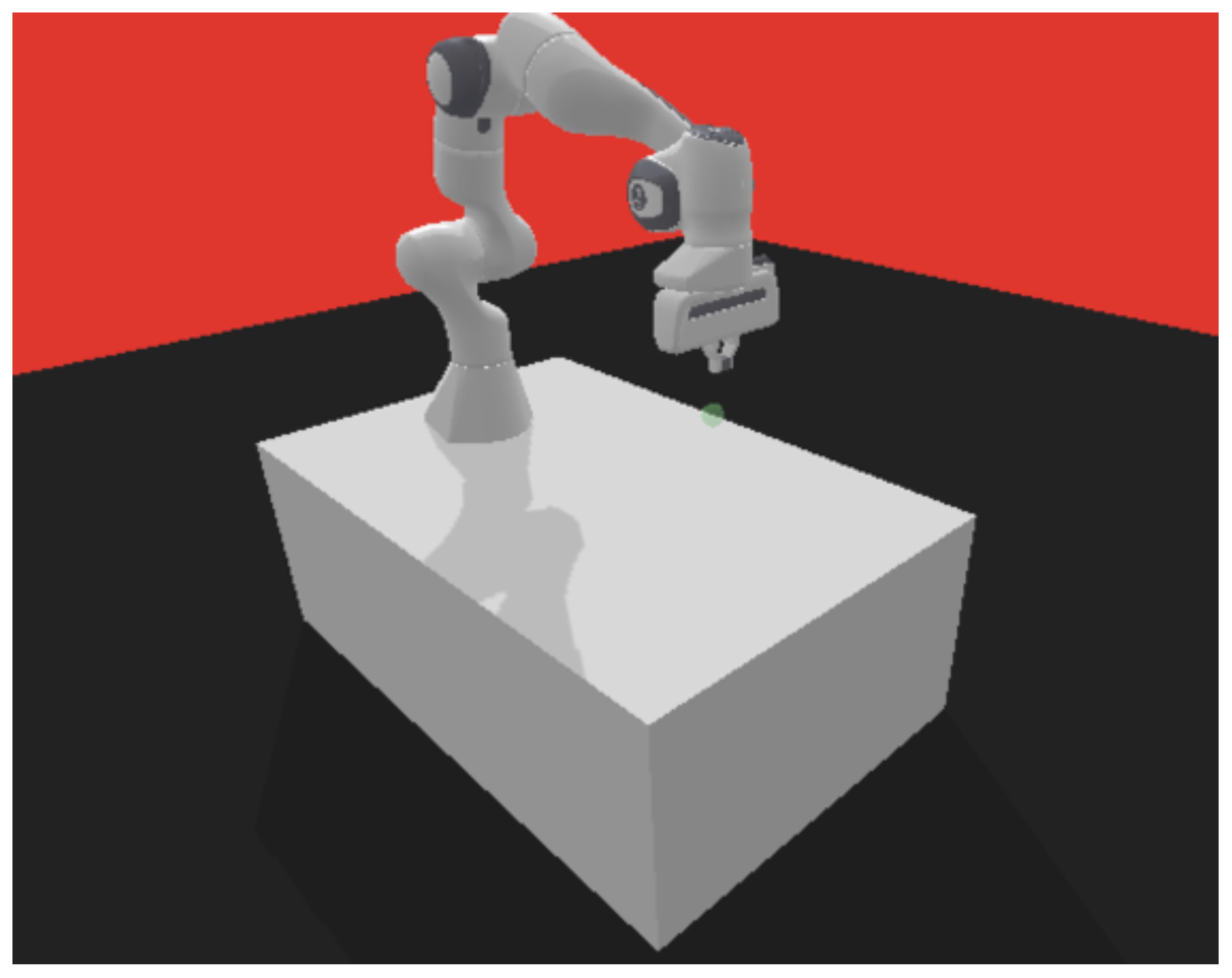}
        \caption{PandaReach-v3}
    \end{subfigure}
    \begin{subfigure}{0.24\textwidth}
        \centering
        \includegraphics[width=\linewidth]{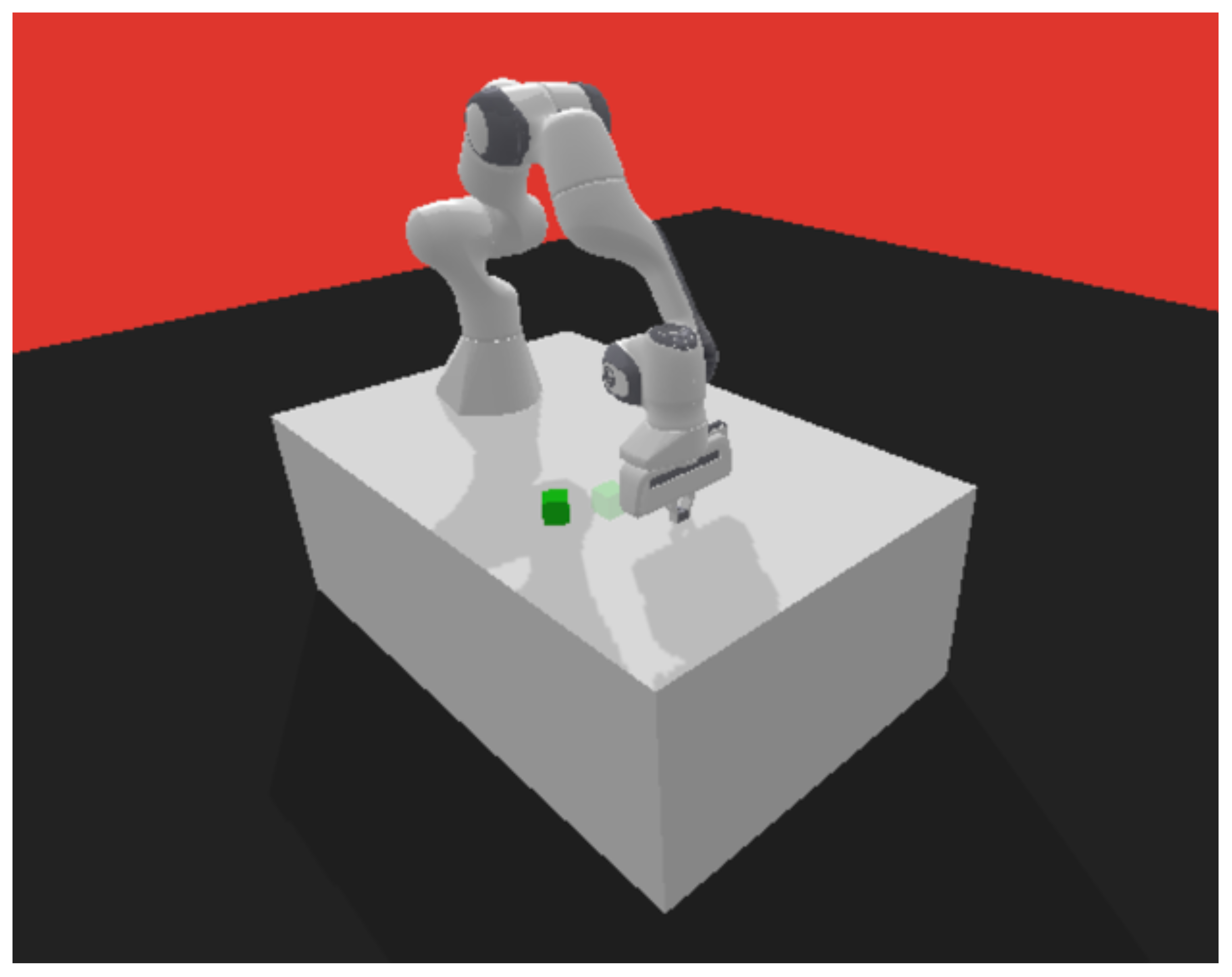}
        \caption{PandaPush-v3}
    \end{subfigure}
    \caption{Benchmark environments in our experiments.}
    \label{fig:Env}
\end{figure*}

We customized several environments based on the D4RL \cite{fu2020d4rl} PointMaze environment. In these custom environments, the agent’s actions correspond to displacement in the x and y directions, instead of applying forces in the x and y axes. We created a set of environments called TrapMaze. In these environments, several traps can cause the agent to get stuck. Once the agent is trapped, it cannot escape regardless of the actions it takes. In the classic U-Maze environment, the goal is represented by a red circle on the map, while the agent is shown as a green circle. The maximum number of timesteps allowed in this environment is 100. In the TrapMaze environments, blue circles represent the traps, the red circle represents the goal, and the green circle represents the agent. We designed two types of TrapMaze environments for our experiments: TrapMaze-v1,which includes a gap in the solid wall that could potentially serve as a shortcut for the agent, and TrapMaze-v2. a more challenging version of TrapMaze-v1, where the shortcut is removed. In this version, if the agent’s reward is based solely on the distance to the goal, it will struggle to perform well. Instead, the agent needs to learn the actual maze layout to successfully reach the goal. Both TrapMaze environments have the maximum number of timesteps set to 300. 

For the robotic manipulation benchmarks, we use the PandaReach and PandaPush environments from panda-gym \cite{gallouedec2021pandagym}. All experimental environments are shown in \cref{fig:Env}.
In our experiments, we also use the PandaPush environment. Due to limited computational resources and to obtain results efficiently, we fixed the object’s position at (0.0, 0.0, 0.02). The goal was restricted to 12 fixed positions, with the goal location for each episode randomly selected from the following set: (0.1, 0., 0.02), (0.12, 0., 0.02), (0.15, 0., 0.02), (0.2, 0., 0.02), (0.1, -0.05, 0.02), (0.12, -0.05, 0.02), (0.15, -0.05, 0.02), (0.2, -0.05, 0.02), (0.1, 0.05, 0.02), (0.12, 0.05, 0.02), (0.15, 0.05, 0.02), (0.2, 0.05, 0.02).

\begin{figure*}[tbh!]
    \centering
    \begin{subfigure}{0.25\textwidth}
        \centering
        \includegraphics[width=\linewidth]{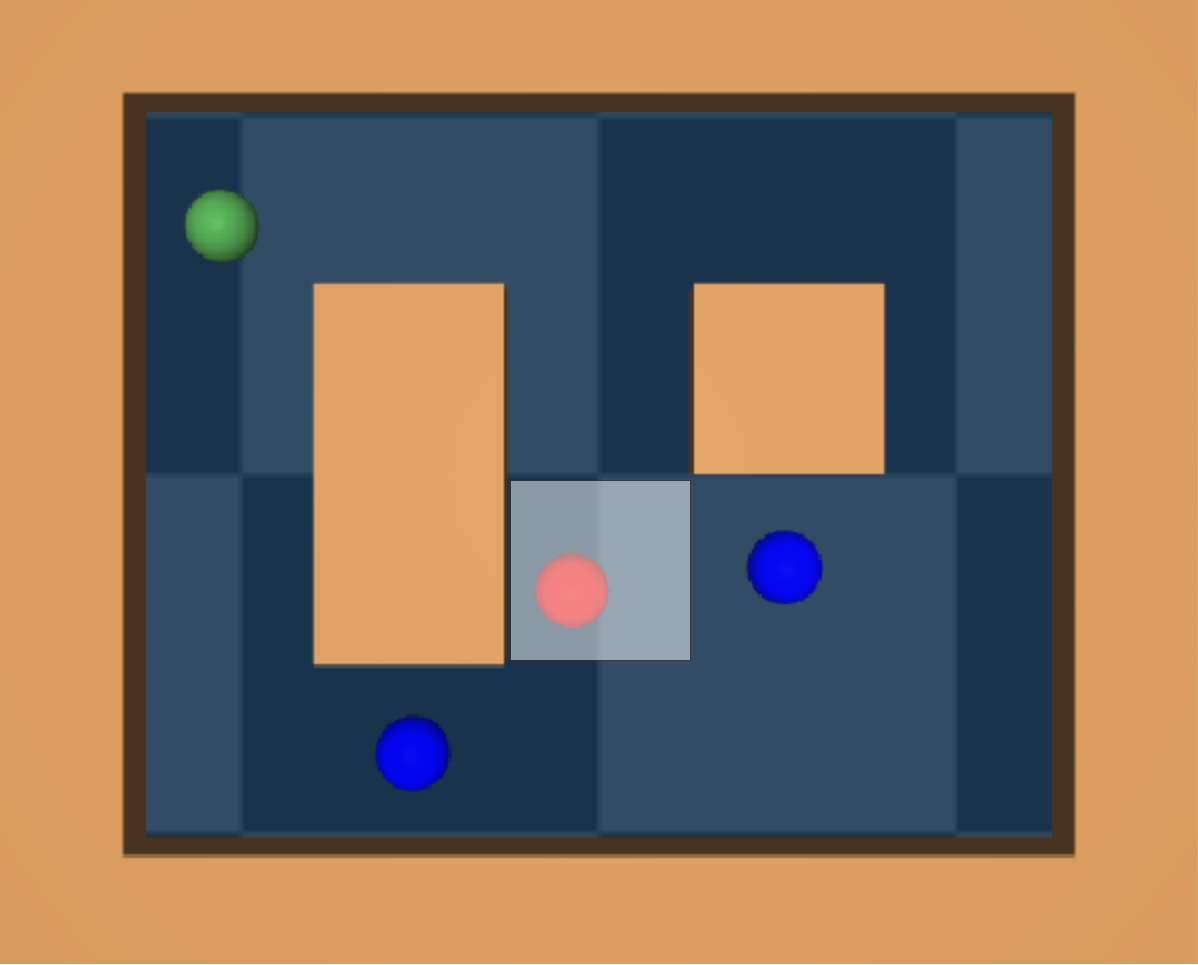}
        \caption{Initialization in 1 grid}
        \label{fig:mul_1}
    \end{subfigure}
    \hspace{0.05\textwidth}
    \begin{subfigure}{0.25\textwidth}
        \centering
        \includegraphics[width=\linewidth]{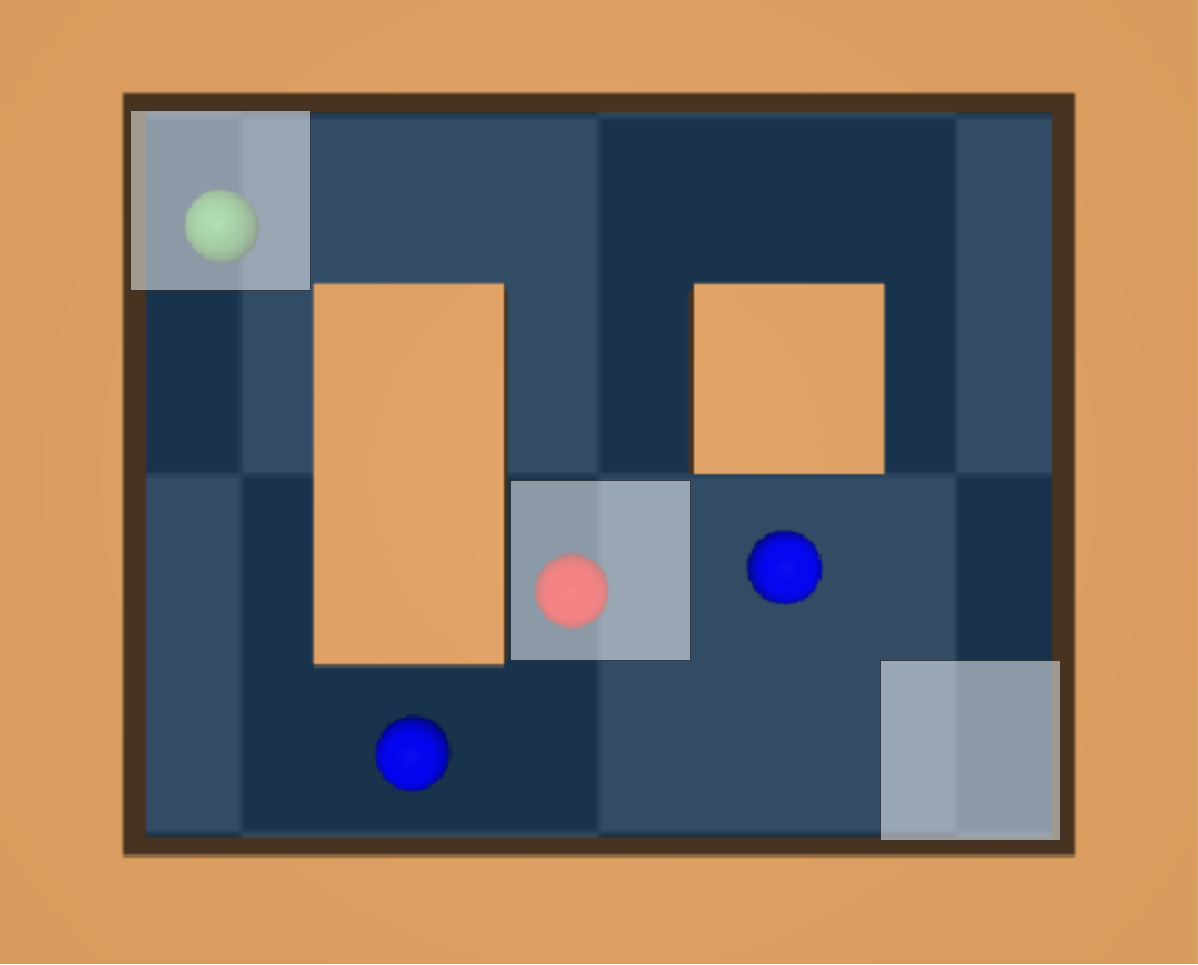}
        \caption{Initialization  in 3 grids}
        \label{fig:mul_3}
    \end{subfigure}
    \hspace{0.05\textwidth}
    \begin{subfigure}{0.25\textwidth}
        \centering
        \includegraphics[width=\linewidth]{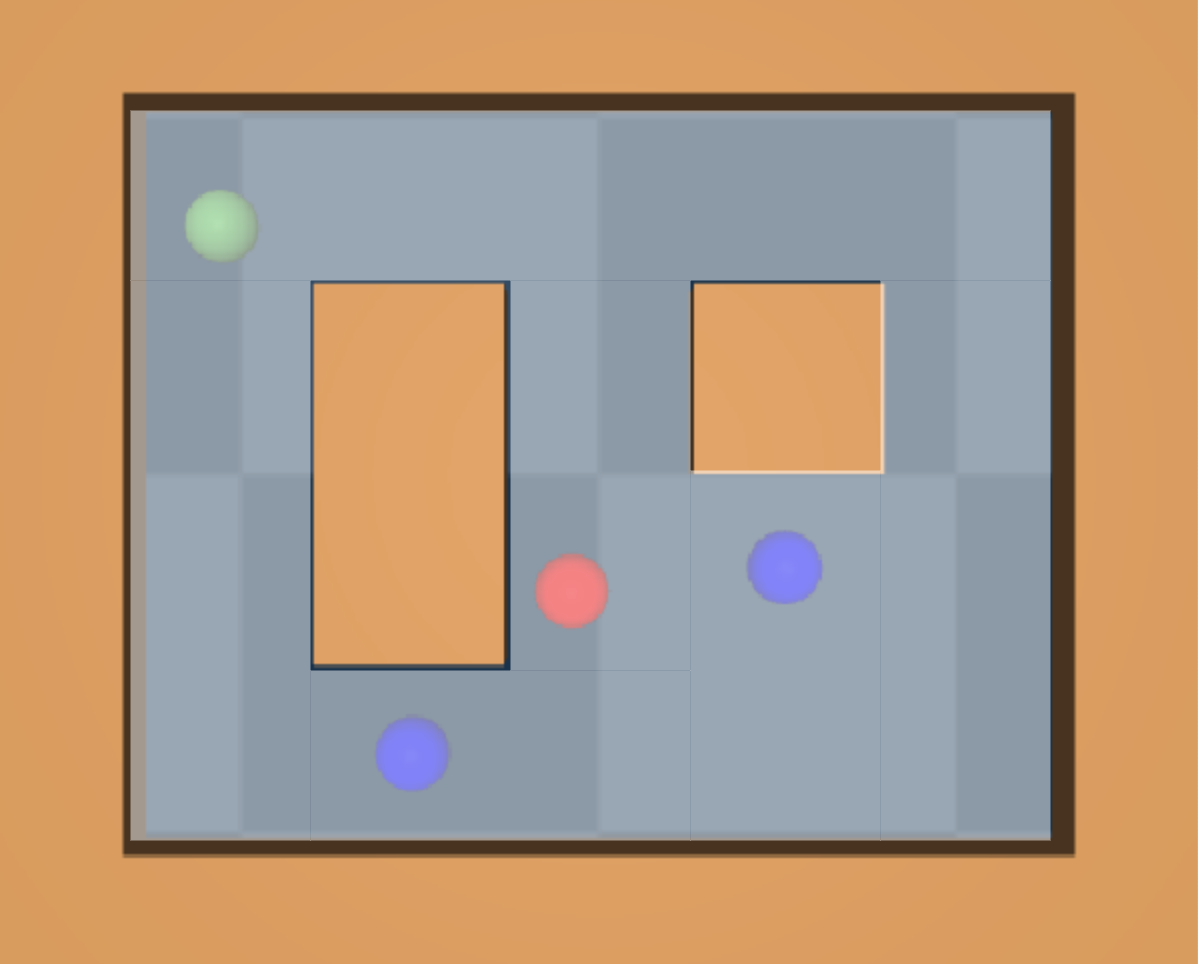}
        \caption{Initialization in any grid}
        \label{fig:mul_all}
    \end{subfigure}
    \caption{Illustration of the Goal Initialization Area.}
    \label{fig:mul}
\end{figure*}

In the PointMaze environments, the goal is randomly initialized within specific grid cells. It cannot be placed in the same grid cell as the agent’s initial position or any traps. In \cref{gen}, we evaluate the performance of TW-CRL under different configurations of initial goal grid placements. In these experiments, the provided demonstrations are only conducted in an environment where the goal is generated in a single grid cell, as shown in \cref{fig:mul_1}. \cref{fig:mul} illustrates these configurations, where the goal can be randomly initialized within the areas shaded in white. Specifically, \cref{fig:mul_1} shows that the goal is initialized in a single specific grid cell, \cref{fig:mul_3} shows the goal initialized in three specific grid cells, and \cref{fig:mul_all} demonstrates the goal initialized in 13 specific grid cells, meaning it can appear in any grid cell within the maze.

\subsection{Experiment Details}\label{ap:hyper}
This section describes the experimental setup and hyperparameters. In our experiments, TW-CRL is trained based on TD3. The hyperparameters used are listed in \cref{tbl:hyperparams}. One thing to note is that our experiment plots cover only a limited range of timesteps. As a result, the end points of the curves in the figures differ from the values in the results table, which report the fully converged results. We limit the plotted range to make differences in trend between methods more visible and to help readers compare them more easily.

In the PandaPush environment, due to limited computational resources and to obtain results more efficiently, we first train an expert policy using the TQC algorithm with Hindsight Experience Replay (HER). This expert is then used to collect successful demonstrations. The actor in our experiments is initially trained with BC using these demonstrations and then further trained using an IRL method.

\begin{table}[tbh!]
\caption{Detailed hyperparameters.}
\label{tbl:hyperparams}
\vskip 0.15in
\begin{center}
\begin{small}
\begin{sc}
\begin{tabular}{lc}
\toprule
\textbf{Hyperparameters} & \textbf{Value} \\
\midrule
\multicolumn{2}{l}{\textit{Shared}} \\
\textit{Replay buffer capacity}    & 1000000  \\
\textit{Batch size}  & 512 \\
\textit{Action bound}    & [-1, 1] \\
\textit{Hidden layers in critic network}    & [256, 256, 256] \\
\textit{Hidden layers in actor network}     & [256, 256, 256] \\
\textit{Optimizer}       & \textit{Adam}  \\
\textit{Actor learning rate}      & $1e-4$        \\
\textit{Critic learning rate}   & $1e-3$  \\
\textit{Discount factor ($\gamma$)}    & 0.99  \\
\textit{Soft update rate ($\tau$)}   & 0.005  \\
\textit{Policy update interval}   & 2  \\
\textit{Start timesteps}   & 100  \\
\textit{Noise std}   &0.2  \\
\textit{Noise clip}   &0.5  \\
\midrule
\multicolumn{2}{l}{\textit{TW-CRL}} \\
\textit{Hidden layers in reward network}     & [128, 128, 128] \\
\textit{Reward learning rate}   & $1e-3$  \\
\textit{Reward update interval}   & 10  \\
\textit{Reward update epoch}   & 200  \\
\midrule
\multicolumn{2}{l}{\textit{MERIT}} \\
\textit{Hidden layers in discriminator network}     & [256, 256, 256] \\
\textit{Discriminator learning rate}   & $1e-3$  \\
\textit{meta regularization hyperparameter ($\lambda_meta$)}   & $0.1$  \\
\midrule
\multicolumn{2}{l}{\textit{ICRL}} \\
\textit{Hidden layers in discriminator network}     & [256, 256, 256] \\
\textit{Discriminator learning rate}   & $1e-3$  \\
\midrule
\multicolumn{2}{l}{\textit{GAIL}} \\
\textit{Hidden layers in discriminator network}     & [256, 256, 256] \\
\textit{Discriminator learning rate}   & $1e-3$  \\
\midrule
\multicolumn{2}{l}{\textit{AIRL}} \\
\textit{Hidden layers in reward network}     & [256, 256, 256] \\
\textit{Reward learning rate}   & $1e-3$  \\
\midrule
\multicolumn{2}{l}{\textit{MaxEnt}} \\
\textit{Hidden layers in reward network}     & [256, 256, 256] \\
\textit{Reward learning rate}   & $3e-4$  \\
\midrule
\multicolumn{2}{l}{\textit{SASR}} \\
\textit{reward weight ($\lambda$)}     & 0.6 \\
\textit{kernel function bandwidth}   & 0.2  \\
\textit{random Fourier features dimension ($\mathcal{M}$)}   & 1000  \\
\textit{retention rate $phi$}   & 0.1  \\
\bottomrule
\end{tabular}
\end{sc}
\end{small}
\end{center}
\vskip -0.1in
\end{table}

\begin{table}[tbh!]
\caption{Detailed hyperparameters for each environment.}
\label{tbl:hyperparams}
\vskip 0.15in
\begin{center}
\begin{small}
\begin{sc}
\begin{tabular}{lc}
\toprule
\textbf{Hyperparameters} & \textbf{Value} \\
\midrule
\multicolumn{2}{l}{\textit{U-Maze}} \\
\textit{Time-Weighted Hyperparameter ($\alpha$)}     & $2$ \\
\textit{Horizon ($T$)}   & $300$  \\
\midrule
\multicolumn{2}{l}{\textit{TrapMaze-v1}} \\
\textit{Time-Weighted Hyperparameter ($\alpha$)}     & $2$ \\
\textit{Horizon ($T$)}   & $300$  \\
\midrule
\multicolumn{2}{l}{\textit{TrapMaze-v2}} \\
\textit{Time-Weighted Hyperparameter ($\alpha$)}     & $2$ \\
\textit{Horizon ($T$)}   & $300$  \\
\midrule
\multicolumn{2}{l}{\textit{MountainCar}} \\
\textit{Time-Weighted Hyperparameter ($\alpha$)}     & $1$ \\
\textit{Horizon ($T$)}   & $999$  \\
\midrule
\multicolumn{2}{l}{\textit{HumanStand}} \\
\textit{Time-Weighted Hyperparameter ($\alpha$)}     & $0.1$ \\
\textit{Horizon ($T$)}   & $1000$  \\
\midrule
\multicolumn{2}{l}{\textit{Ant}} \\
\textit{Time-Weighted Hyperparameter ($\alpha$)}     & $0.1$ \\
\textit{Horizon ($T$)}   & $1000$  \\
\midrule
\multicolumn{2}{l}{\textit{PandaReach}} \\
\textit{Time-Weighted Hyperparameter ($\alpha$)}     & $2$ \\
\textit{Horizon ($T$)}   & $50$  \\
\midrule
\multicolumn{2}{l}{\textit{PandaPush}} \\
\textit{Time-Weighted Hyperparameter ($\alpha$)}     & $2$ \\
\textit{Horizon ($T$)}   & $100$  \\
\bottomrule
\end{tabular}
\end{sc}
\end{small}
\end{center}
\vskip -0.1in
\end{table}

\clearpage
\subsection{Hardware Information}\label{ap:resource}
We conducted our experiments on a GPU server. On 1 Nvidia L40 GPU and 8 Intel Xeon Platinum 8462Y CPUs, each run of TW-CRL takes around 4 to 24 hours depending on the complexity of the environment. 

\subsection{License Information}\label{ap:license}
Our paper uses existing open-source libraries for our environments and baseline methods. All assets are under permissive licenses such as MIT and CC-BY-4.0, and we have made proper citations.

\section{Additional Results} \label{ap:experiment}

\subsection{Additional Results for Generalization}\label{ap:ad-generalization}

\begin{figure*}[h]
    \centering
    \begin{subfigure}{0.29\textwidth}
        \centering
        \includegraphics[width=\linewidth]{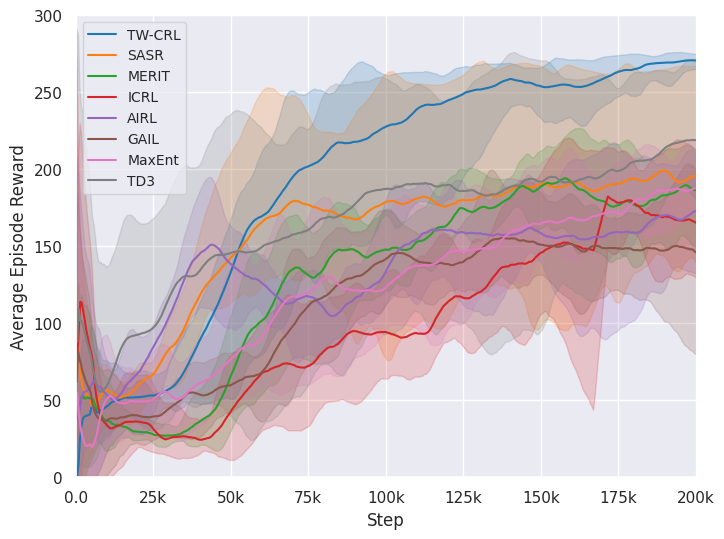}
    \end{subfigure}
    \begin{subfigure}{0.29\textwidth}
        \centering
        \includegraphics[width=\linewidth]{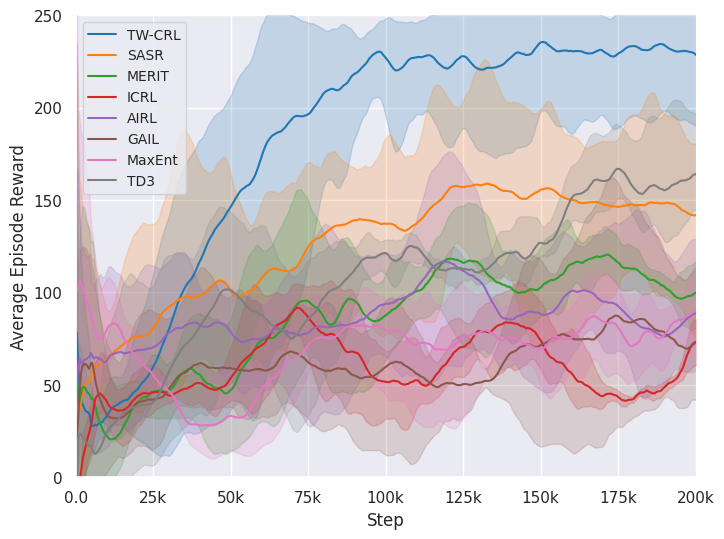}
    \end{subfigure}
    \begin{subfigure}{0.29\textwidth}
        \centering
        \includegraphics[width=\linewidth]{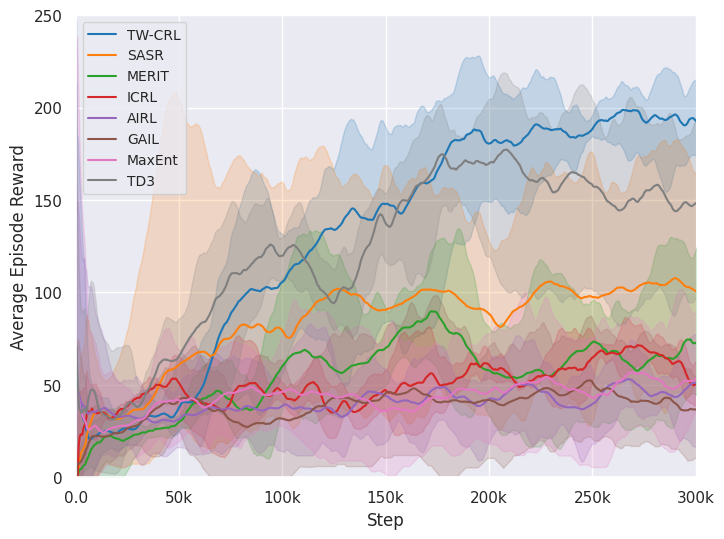}
    \end{subfigure}

    \begin{subfigure}{0.29\textwidth}
        \centering
        \includegraphics[width=\linewidth]{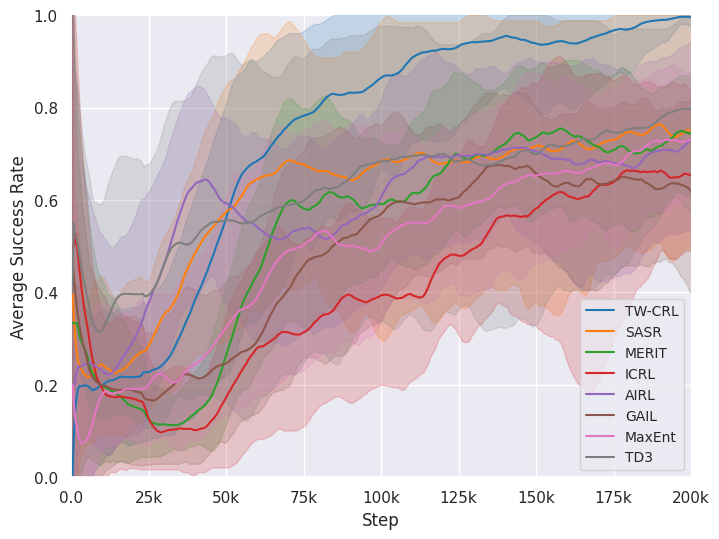}
        \caption{Initialization in 1 grid }
    \end{subfigure}
    \begin{subfigure}{0.29\textwidth}
        \centering
        \includegraphics[width=\linewidth]{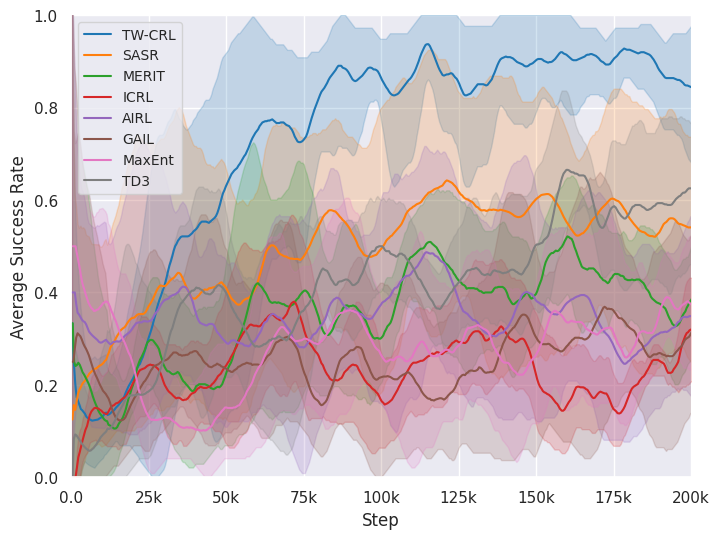}
        \caption{Initialization in 3 grids }
    \end{subfigure}
    \begin{subfigure}{0.29\textwidth}
        \centering
        \includegraphics[width=\linewidth]{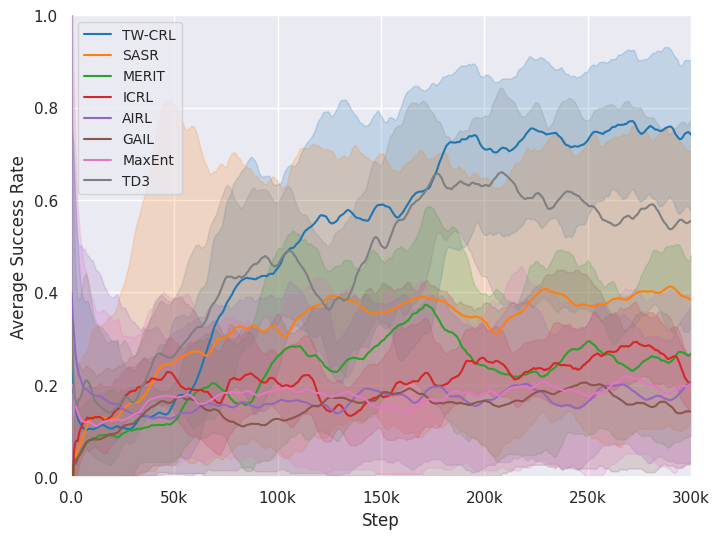}
        \caption{Initialization in any grid }
    \end{subfigure}
    \caption{Training and success rate curves on TrapMaze-v1 under three different configurations. The solid curves represent the mean, and the shaded regions indicate the standard deviation over five runs. The first row shows the average return curves for each setting, while the second row shows the}
    \label{fig:mul_plot}
\end{figure*}

We provide additional results from our generalization experiments here. \cref{fig:mul_plot} shows the average episode reward and success rate curves on TrapMaze-v1 under three different configurations mentioned in \cref{gen}. \cref{tab:multi_goal} shows the average episode reward of other tested baseline methods not shown in \cref{tab:multi_goal} due to page limit.
\begin{table}[h!]
\centering
\caption{Additional average return under three goal reset settings. 1, 3, and any. \#goal represents the number of grids where the goal state might be in.}
\footnotesize
\begin{tabular}{cccccc}
\toprule
\text{\# goal state(s)}    & \text{TW-CRL}                           & \text{TD3}             & \text{MaxEnt}             & \text{AIRL}                & \text{GAIL}                          \\ \midrule
\multirow{1}{*}{1}
                         & \textbf{253.12} ± \textbf{9.46}          & 196.28 ± 32.89          & 129.70 ± 18.50      & 160.22± 11.25           & 159.06 ± 25.53 \\  
\multirow{1}{*}{3} 
                         & \textbf{240.48 } ± \textbf{4.44}          & 168.58 ± 27.86           & 73.24 ± 20.10      & 74.26 ± 20.81          & 59.79 ± 22.87 \\
\multirow{1}{*}{Any} 
                         & \textbf{221.33} ± \textbf{11.58}         & 168.38 ± 40.06           & 49.77 ± 33.81     & 57.55 ± 30.27            & 55.25 ± 21.67 \\

\bottomrule
\end{tabular}
\label{tab:multi_goal_2}
\end{table}

\subsection{Additional Results for Reward Function Visualizations}\label{ap:ad-reward}
We provide additional results of visualization of reward function in TrapMaze-v1 environment in \cref{fig:traj}.
\begin{figure*}[h!]
    \centering
    \includegraphics[width=\linewidth]{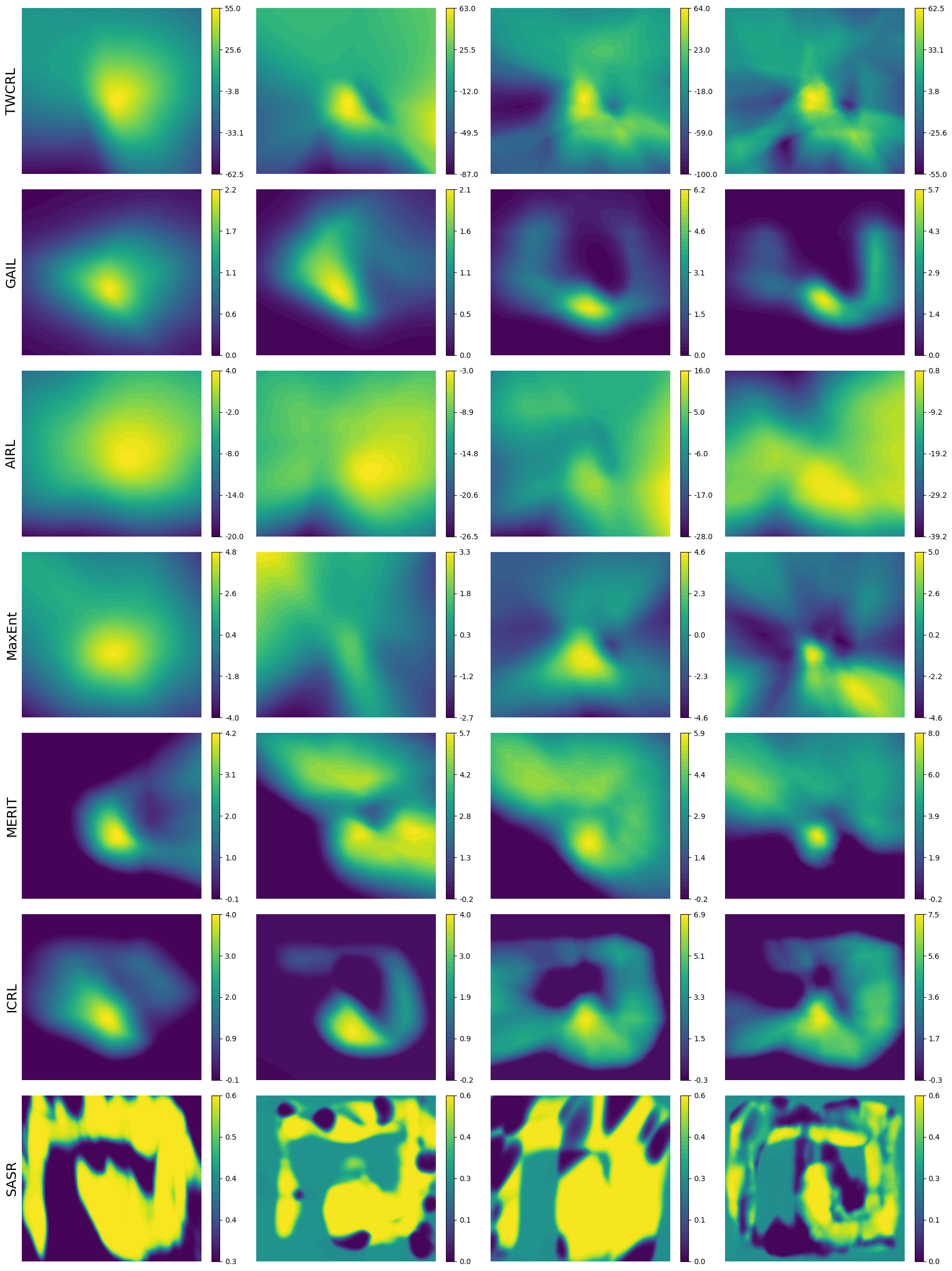}
    \caption{Visualization of the reward function in TrapMaze-v1.}
    \label{fig:traj}
\end{figure*}

\section{Limitation and Future Work}
In this paper, we introduce TW‑CRL as a new IRL method that uses both successful and failed demonstrations along with the temporal information of each state to learn a better reward function and train an agent’s policy. In its current form, TW‑CRL works best on tasks where a series of actions unfolds over time, such as navigation problems. It is not easily applicable by design for tasks that require highly repetitive or back‑and‑forth actions, like Humanoid or the HalfCheetah. Its performance also depends a lot on how the Time‑Weighted function and the success threshold are set. With careful tuning of these hyperparameters, TW‑CRL can be adapted to other environments, but this tuning must be done separately for each new task.

In future work, we will explore different forms of the time‑weight function and test how they affect learning. We will also investigate using TW‑CRL for reward shaping by adding the learned reward to the environment’s own reward, which may speed up training on tasks that already provide clear feedback. Finally, we aim to develop simple methods for automatically selecting the Time-Weighted function and success threshold so that TW‑CRL can be applied more robustly across a wider range of environments.

\end{document}